%% file: main.tex
\documentclass{article}
\PassOptionsToPackage{square,comma,numbers,compress}{natbib}

\usepackage[preprint]{neurips_2023}

\usepackage[utf8]{inputenc} %
\usepackage[T1]{fontenc}    %
\usepackage[hidelinks,colorlinks=true]{hyperref}       %
\usepackage[dvipsnames]{xcolor}
\usepackage{url}            %
\usepackage{booktabs}       %
\usepackage{amsfonts}       %
\usepackage{nicefrac}       %
\usepackage{microtype}      %
\usepackage{xcolor}         %
\usepackage{graphicx}
\graphicspath{ {figures/} }
\usepackage{array}
\usepackage{times,graphicx,tabulary,multirow,xspace}
\usepackage{subfig}
\usepackage{changepage}
\usepackage{color, colortbl}

\makeatletter
\DeclareRobustCommand\onedot{\futurelet\@let@token\@onedot}
\def\@onedot{\ifx\@let@token.\else.\null\fi\xspace}

\def\eg{\emph{e.g}\onedot}

\def\etal{\emph{et al}\onedot}
\makeatother

\newcommand{\modelname}{GPT-4V\xspace}
\newcommand{\modelnamefull}{GPT-4V(ision)\xspace}

\title{On the Road with \modelnamefull: Early Explorations of Visual-Language Model on Autonomous Driving}

\author{
{\bf Licheng Wen$^{1\dagger}$, Xuemeng Yang$^{1\dagger}$, Daocheng Fu$^{1\dagger}$, Xiaofeng Wang$^{2\dagger}$, Pinlong Cai$^{1}$, Xin Li$^{1,3}$,} \\ \\
{\bf Tao Ma$^{1,4}$, Yingxuan Li$^{2}$, Linran Xu$^{2}$, Dengke Shang$^{2}$, Zheng Zhu$^{2\spadesuit}$, Shaoyan Sun$^{2}$,} \\ \\
{\bf Yeqi Bai$^{1}$, Xinyu Cai$^{1}$, Min Dou$^{1}$, Shuanglu Hu$^{5}$, Botian Shi$^{1\spadesuit}$, Yu Qiao$^{1}$} \\ \\
$^\dagger$~Core Contributors \; 
$^{\spadesuit}$~Corresponding Authors\\ \\
$^{1}$~Shanghai Artificial Intelligence Laboratory, Shanghai, China \\ \\
$^{2}$~GigaAI, Beijing, China \;
$^{3}$~East China Normal University, Shanghai, China \\ \\
$^{4}$~The Chinese University of Hong Kong, Hong Kong, China \;
$^{5}$~WeRide.ai, Shanghai, China \\ \\
\texttt{shibotian@pjlab.org.cn}~~~~\texttt{zhengzhu@ieee.org} \\ \\
}

\begin{document}

\maketitle

\begin{abstract}
The pursuit of autonomous driving technology hinges on the sophisticated integration of perception, decision-making, and control systems. Traditional approaches, both data-driven and rule-based, have been hindered by their inability to grasp the nuance of complex driving environments and the intentions of other road users. This has been a significant bottleneck, particularly in the development of common sense reasoning and nuanced scene understanding necessary for safe and reliable autonomous driving.
The advent of Visual Language Models (VLM) represents a novel frontier in realizing fully autonomous vehicle driving. This report provides an exhaustive evaluation of the latest state-of-the-art VLM, \modelnamefull, and its application in autonomous driving scenarios. We explore the model's abilities to understand and reason about driving scenes, make decisions, and ultimately act in the capacity of a driver. Our comprehensive tests span from basic scene recognition to complex causal reasoning and real-time decision-making under varying conditions.
Our findings reveal that \modelname demonstrates superior performance in scene understanding and causal reasoning compared to existing autonomous systems. It showcases the potential to handle out-of-distribution scenarios, recognize intentions, and make informed decisions in real driving contexts. However, challenges remain, particularly in direction discernment, traffic light recognition, vision grounding, and spatial reasoning tasks. These limitations underscore the need for further research and development.
Project is now available on GitHub for interested parties to access and utilize: \url{https://github.com/PJLab-ADG/GPT4V-AD-Exploration}
\end{abstract}
\vspace{20pt}

\clearpage
{
  \hypersetup{linkcolor=black}
  \tableofcontents
  \label{sec:toc}
}

\clearpage
{
\hypersetup{linkcolor=black}
\addcontentsline{toc}{section}{List of Figures}
\listoffigures
\label{sec:lof}
}
\clearpage

\input{sections/intro.tex}
\input{sections/scenario-understanding.tex}

\input{sections/reasoning.tex}
\input{sections/driver-agent.tex}
\input{sections/limitations-conclusions.tex}

\clearpage
{
\bibliographystyle{plain}
\bibliography{egbib}
}

\end{document}

%% file: sections/intro.tex
\begin{figure}[tbp]
    \centering
    \includegraphics[width=\textwidth]{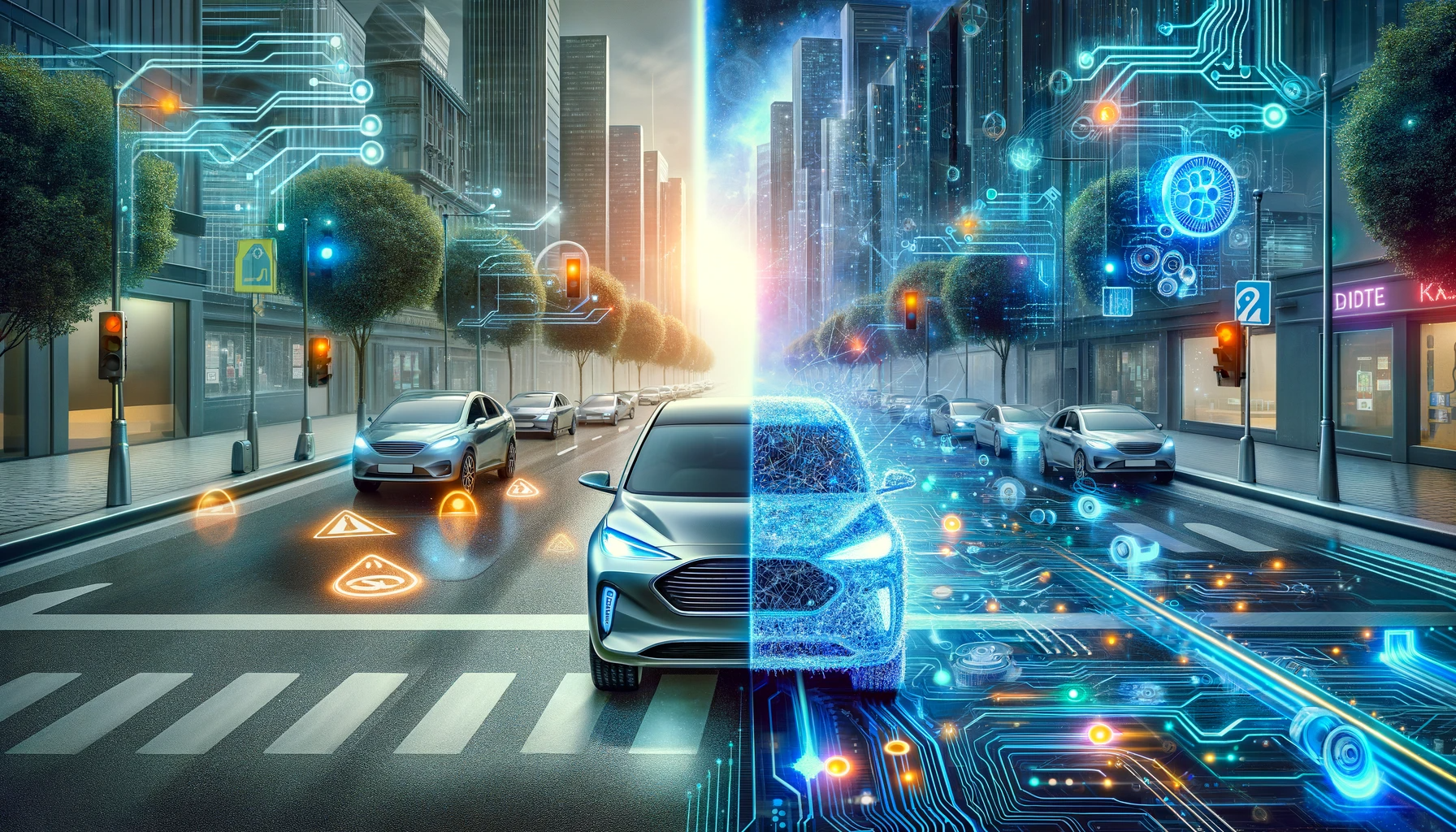}
    \caption[Section \ref{sec:01intro}: Introduction figure]{An illustration showing the transition from the traditional autonomous driving pipeline to the integration of visual language models such as GPT-4V. This picture is generated by DALL·E 3.}
    \label{fig: intro}
\end{figure}

\section{Introduction}
\label{sec:01intro}
\subsection{Motivation and Overview}

The quest for fully autonomous vehicles has long been constrained by a pipeline that relies on perception, decision-making, and planning control systems. Traditional approaches, whether rooted in data-driven algorithms or rule-based methods, fall short in several key areas. Specifically, they exhibit weaknesses in accurately perceiving open-vocabulary objects and struggle with interpreting the behavioral intentions of surrounding traffic participants. 
The reason is that traditional approaches only characterize abstract features of limited acquisition data or deal with problems according to predetermined rules, whereas they lack the ``common sense'' to deal with rare but significant corner cases and fail to summarize driving-related knowledge from the data for nuanced scenario understanding and effective causal reasoning.

The emergence of Large Language Models (LLMs), exemplified by GPT-3.5~\cite{gpt3.5}, GLM~\cite{du2022glm, zeng2022glm},  Llama~\cite{touvron2023llama, touvron2023llama2}, \etal, has shown a glimmer of hope in addressing these issues. The LLMs are equipped with a rudimentary form of common sense reasoning, thereby showing promise in understanding complex driving scenarios. However, their application in autonomous driving has been restricted mainly to decision-making and planning phases~\cite{fu2023drive, wen2023dilu,chen2023driving, mao2023gpt}. This limitation is due to their inherent inability to process and understand visual data, which is critical for accurately perceiving the driving environment and driving the vehicle safely.

The recent development of \modelname~\cite{gpt4v, gpt4v_2, gpt4v_3,yang2023dawn}, a cutting-edge Vision-Language Model (VLM), has opened up new vistas for research and development. Unlike its predecessors (GPT-4~\cite{gpt4}), \modelname possesses robust capabilities in image understanding, marking a significant step forward in closing the perception gap in autonomous driving technologies. This newfound strength raises the question: Can \modelname serve as a cornerstone for improving scene understanding and causal reasoning in autonomous driving?

In this paper, we aim to answer this pivotal question by conducting an exhaustive evaluation of \modelname's abilities. Our research delves into the model's performance in the intricate aspects of scene understanding and causal reasoning within the domain of autonomous driving. Through exhaustive testing and in-depth analysis, we have elucidated both the capabilities and limitations of \modelname, which is anticipated to offer valuable support for researchers to venture into potential future applications within the autonomous driving industry.

We have tested the capabilities of \modelname with increasing difficulty, from scenario understanding to reasoning, and finally testing its continuous judgment and decision-making ability as drivers in real-world driving scenarios. Our exploration of \modelname in the field of autonomous driving mainly focuses on the following aspects:
\begin{enumerate}
    \item \textbf{Scenario Understanding:} This test aims to assess \modelname's fundamental recognition abilities. It involves recognizing weather and illumination conditions while driving, identifying traffic lights and signs in various countries, and assessing the positions and actions of other traffic participants in photos taken by different types of cameras. Additionally, we explored simulation images and point cloud images of different perspectives for curiosity's sake.
    \item \textbf{Reasoning:} In this phase of the test, we delve deeper into assessing \modelname's causal reasoning abilities within autonomous driving contexts. This evaluation encompasses several crucial aspects. Firstly, we scrutinize its performance in tackling complex corner cases, which often challenge data-driven perception systems. Secondly, we assess its competence in providing a surround view, which is a vital feature in autonomous driving applications. Given \modelname's inability to directly process video data, we utilize concatenated time series images as input to gauge its temporal correlation capabilities. Additionally, we conduct tests to validate its capacity to associate real-world scenes with navigation images, further examining its holistic understanding of autonomous driving scenarios.
    \item \textbf{Act as a driver: } To harness the full potential of \modelname, we entrusted it with the role of a seasoned driver, tasking it with making decisions in real driving situations based on the environment. Our approach involved sampling driving video at a consistent frame rate and feeding it to \modelname frame by frame. To aid its decision-making, we supplied essential vehicle speed and other relevant information and communicated the driving objective for each video. We challenged \modelname to produce the necessary actions and provide explanations for its choices, thereby pushing the boundaries of its capabilities in real-world driving scenarios.
\end{enumerate}

In conclusion, we offer initial insights as a foundation for inspiring future research endeavors in the realm of autonomous driving with \modelname. Building upon the information presented above, we methodically structure and showcase the qualitative results of our investigation using a unique and engaging compilation of image-text pairs. While this methodology may be somewhat less stringent, it affords the opportunity for a comprehensive analysis.

\subsection{Guidance}

This article focuses on testing in the field of autonomous driving, employing a curated selection of images and videos representing diverse driving scenarios. The test samples are sourced from various outlets, including open-source datasets such as nuScenes~\cite{caesar2020nuscenes}, Waymo Open dataset~\cite{sun2020scalability}, Berkeley Deep Drive-X (eXplanation) Dataset (BDD-X)~\cite{kim2018textual}, D$^2$-city~\cite{che2019d}, Car Crash Dataset (CCD)~\cite{BaoMM2020}, TSDD~\cite{TSDD}, CODA~\cite{li2022coda}, ADD~\cite{wu2023add}, as well as V2X datasets like DAIR-V2X~\cite{yu2022dair} and CitySim~\cite{zheng2022citysim}. Additionally, some samples are derived from the CARLA~\cite{dosovitskiy2017carla} simulation environment, and others are obtained from the internet. It's worth noting that the image data used in testing may include images with timestamps up to April 2023, potentially overlapping with the \modelname model's training data, while the text queries employed in this article are entirely generated anew.

All experiments detailed in this paper were conducted before November 5th, 2023, utilizing the web-hosted \modelnamefull (version from September 25th). We acknowledge that the most recent version of \modelname, which has received updates following the November 6th OpenAI DevDay, may produce different responses when presented with the same images compared to our test results.

%% file: sections/scenario-understanding.tex
\clearpage
\section{Basic Capability of Scenario Understanding}
\label{sec:scenarioUnderstanding}

To achieve safe and effective autonomous driving, a fundamental prerequisite is a thorough understanding of the current scenario. Complex traffic scenarios encompass a myriad of driving conditions, each hosting a diverse array of traffic participants. Accurate recognition and comprehension of these elements serve as basic capabilities for an autonomous vehicle to make informed and appropriate driving decisions.
In this section, we present a series of tests aimed at evaluating \modelname's ability to comprehend traffic scenarios. We focus on two primary aspects: the model's understanding of the surrounding environment and its capacity to discern the behavior and status of various traffic participants. Through these assessments, we aim to shed light on \modelname's competence in interpreting the dynamic traffic environment.

\subsection{Understanding of Environment}
\label{sec:environmentUnderstanding}

In assessing the capabilities of \modelname to comprehend its surrounding environments, we conducted a series of tests encompassing the following key aspects: its ability to discern the time of day, its understanding of prevailing weather conditions, and its proficiency in recognizing and interpreting traffic lights and signs. These elements hold paramount significance in shaping the autonomous driving system's decision-making process. For instance, it is evident that driving at night or in challenging weather conditions requires a heightened level of caution, whereas during daylight hours or in favorable weather conditions a more leisurely driving strategy can be adopted.
Besides, the correct interpretation of traffic lights and road signs is essential for the effectiveness of autonomous driving systems.
We utilize vehicles' front-view images as the primary visual input throughout this section. 
The visual data employed here is drawn from nuScenes~\cite{caesar2020nuscenes}, D$^2$-city\cite{che2019d}, BDD-X~\cite{kim2018textual} and TSDD~\cite{TSDD}.

\begin{figure*}[!b]
\centering
\includegraphics[width=\textwidth]{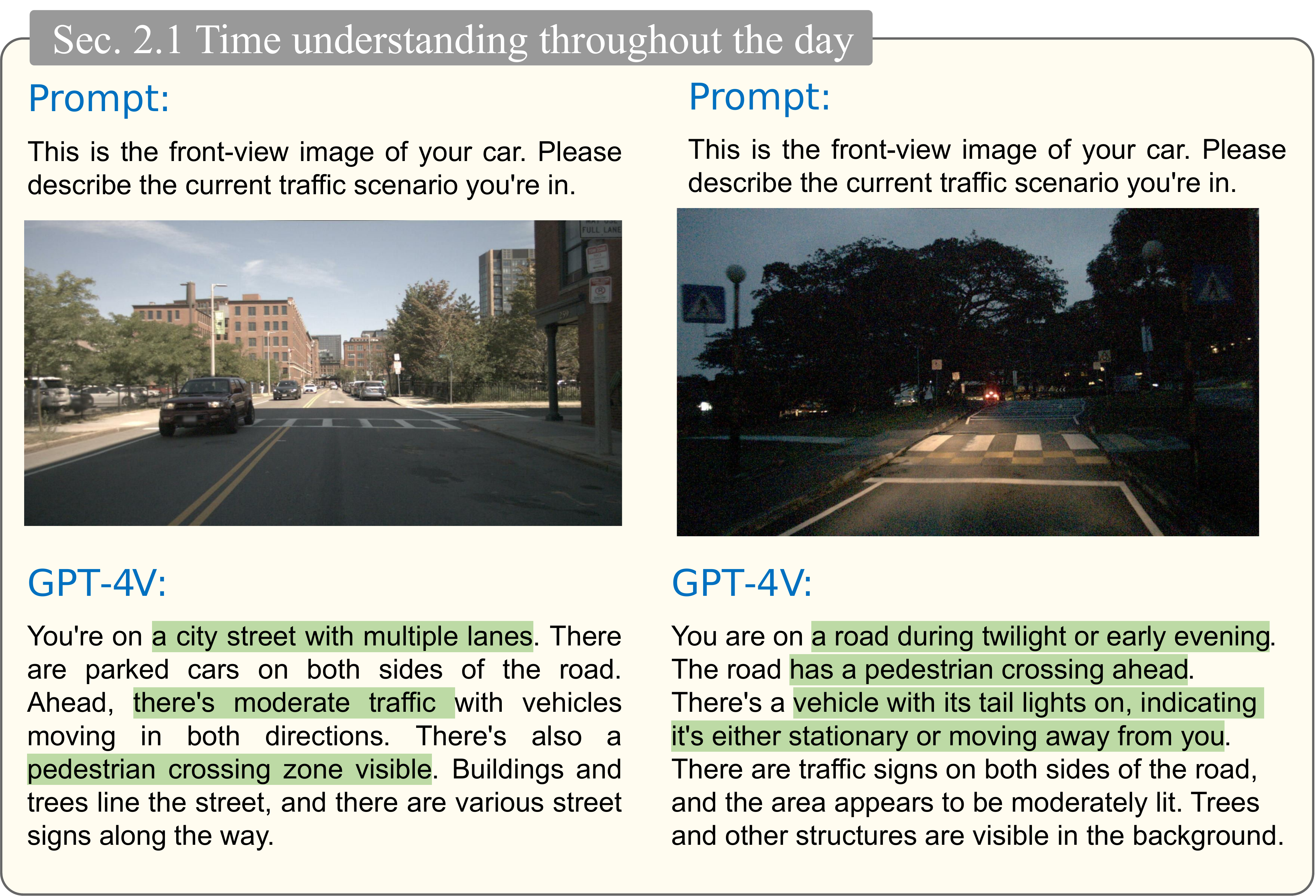}
\caption[Section~\ref{sec:environmentUnderstanding}: Time understanding throughout the day]{ Results on the ability to comprehensively understand time over the course of a day. \colorbox[RGB]{189,218,165}{Green} highlights the right answer in understanding. Check Section~\ref{sec:environmentUnderstanding} for detailed discussions.
}
\label{fig: sec2.1 time understanding}
\end{figure*}

\textbf{Time understanding throughout the day.}
We evaluate \modelname's ability to comprehend temporal differences by providing it with both daytime and nighttime images. We instruct the model to describe the traffic scenarios depicted in these images, and the results are presented in Figure~\ref{fig: sec2.1 time understanding}.
The findings reveal that, when presented with daytime scenes, \modelname successfully identifies them as multi-lane urban roads with ``moderate traffic''. Furthermore, the model adeptly recognizes the presence of a crosswalk on the road.
When confronted with similar nighttime scenes, \modelname's performance is even better. It not only discerns the time as ``twilight or early evening'' but also detects a vehicle with its tail lights on in the distance, and infers that ``it's either stationary or moving away from you''.

\begin{figure*}[htbp]
\centering
\includegraphics[width=\textwidth]{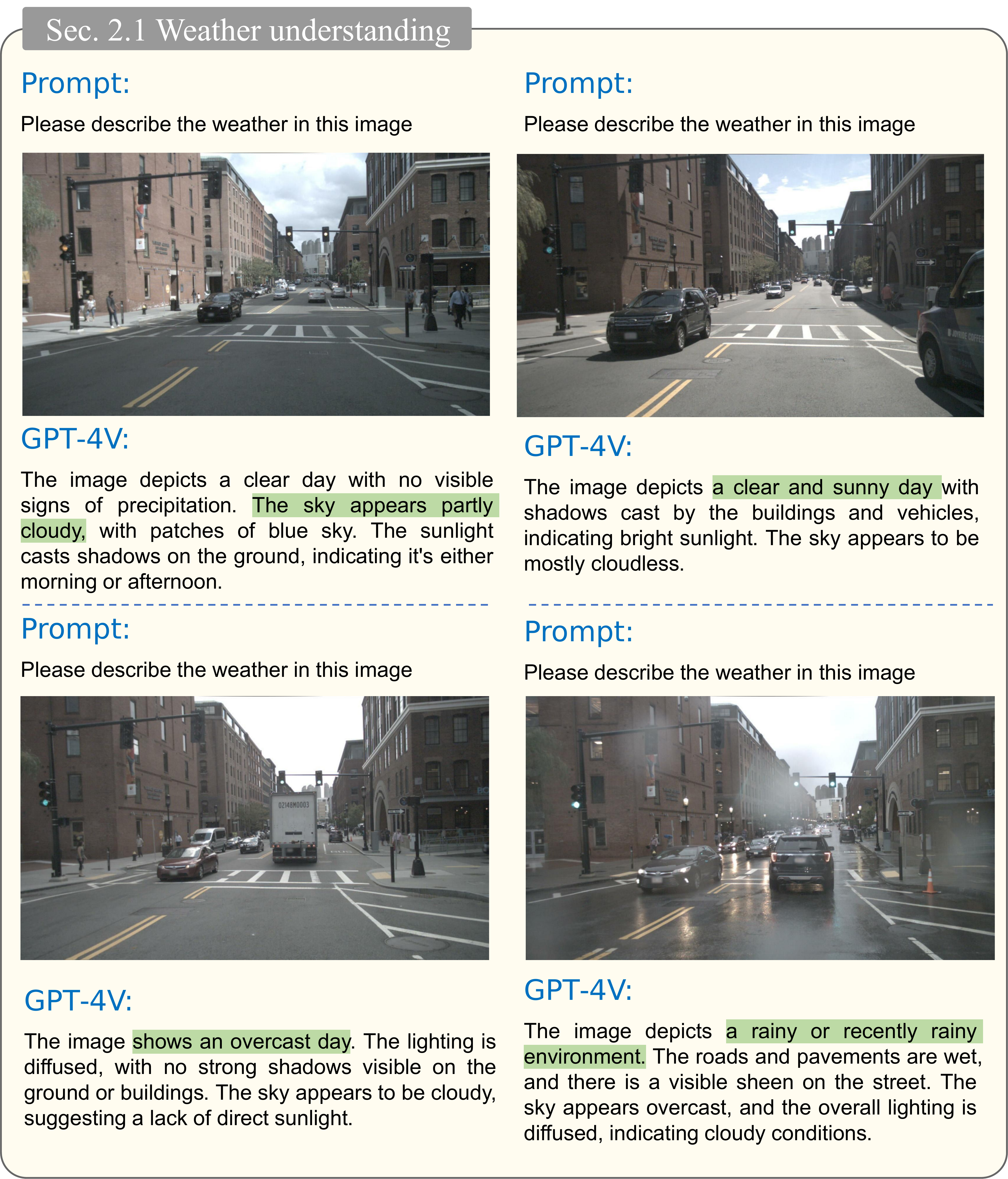}
\caption[Section~\ref{sec:environmentUnderstanding}: Weather understanding]{ Results that identify the season in which the vehicle is driven. \colorbox[RGB]{189,218,165}{Green} highlights the right answer in understanding. Check Section~\ref{sec:environmentUnderstanding} for detailed discussions.
}
\label{fig: sec2.1 weather understanding}
\end{figure*}

\textbf{Weather understanding.} 
Weather is a crucial environmental factor that significantly influences driving behavior. We selected four photographs captured at the same intersection under varying weather conditions from the nuScenes~\cite{caesar2020nuscenes} dataset. We tasked \modelname with identifying the weather conditions depicted in these images. The results are presented in Figure~\ref{fig: sec2.1 weather understanding}.
The results demonstrate that \modelname exhibits remarkable accuracy in recognizing the weather conditions in each image, namely, cloudy, sunny, overcast, and rainy. Moreover, it provides sound justifications for these conclusions, citing factors such as the presence of sunny shadows or the wetness of the streets.

\begin{figure}[htbp]
    \centering
    \includegraphics[width=\textwidth]{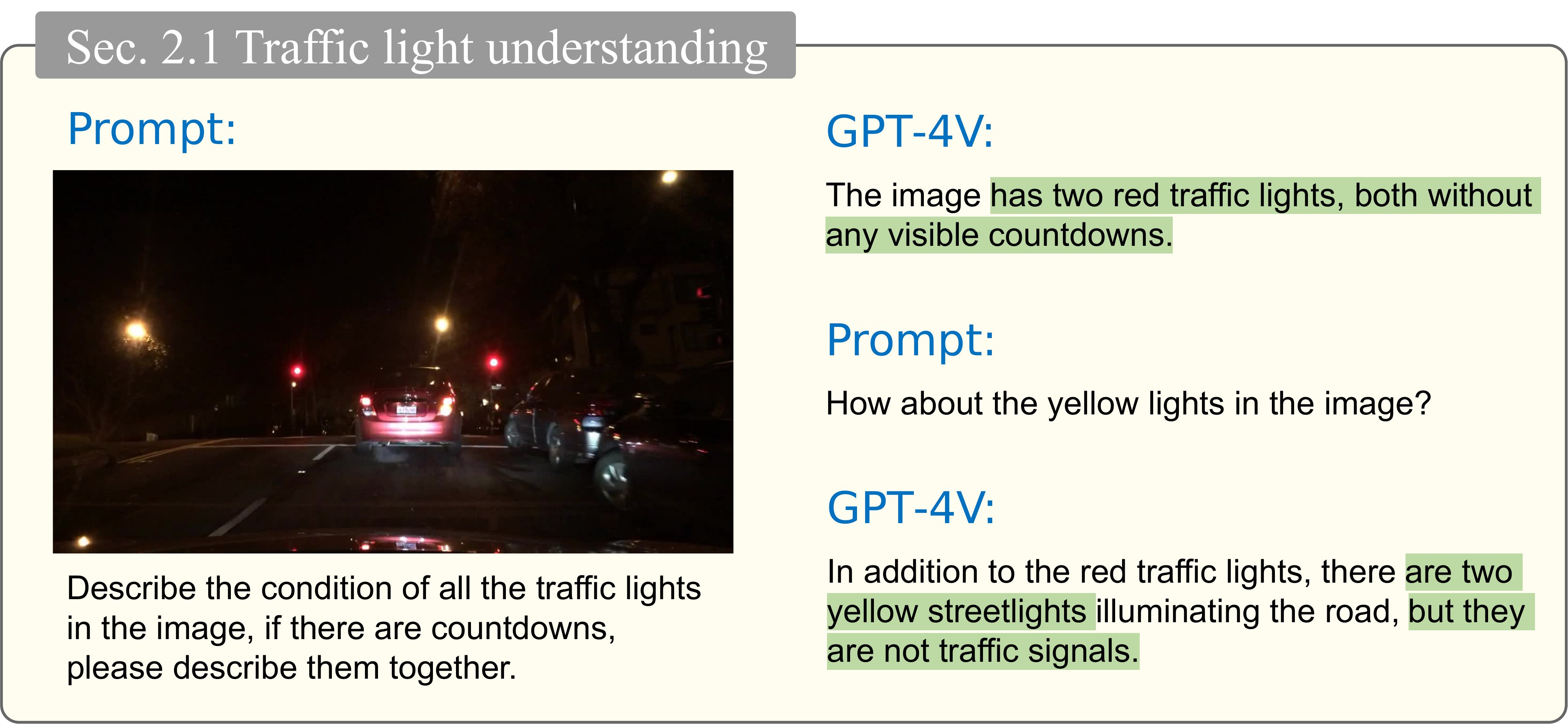}
    \caption[Section~\ref{sec:environmentUnderstanding}: Traffic light understanding, Part I]{Results on the ability to understand the traffic light. \colorbox[RGB]{189,218,165}{Green} highlights the right answer in understanding. Check Section~\ref{sec:environmentUnderstanding} for detailed discussions.}
    \label{fig: sec2.1 traffic light understanding 1}
\end{figure}

\begin{figure}[htbp]
    \centering
    \includegraphics[width=\textwidth]{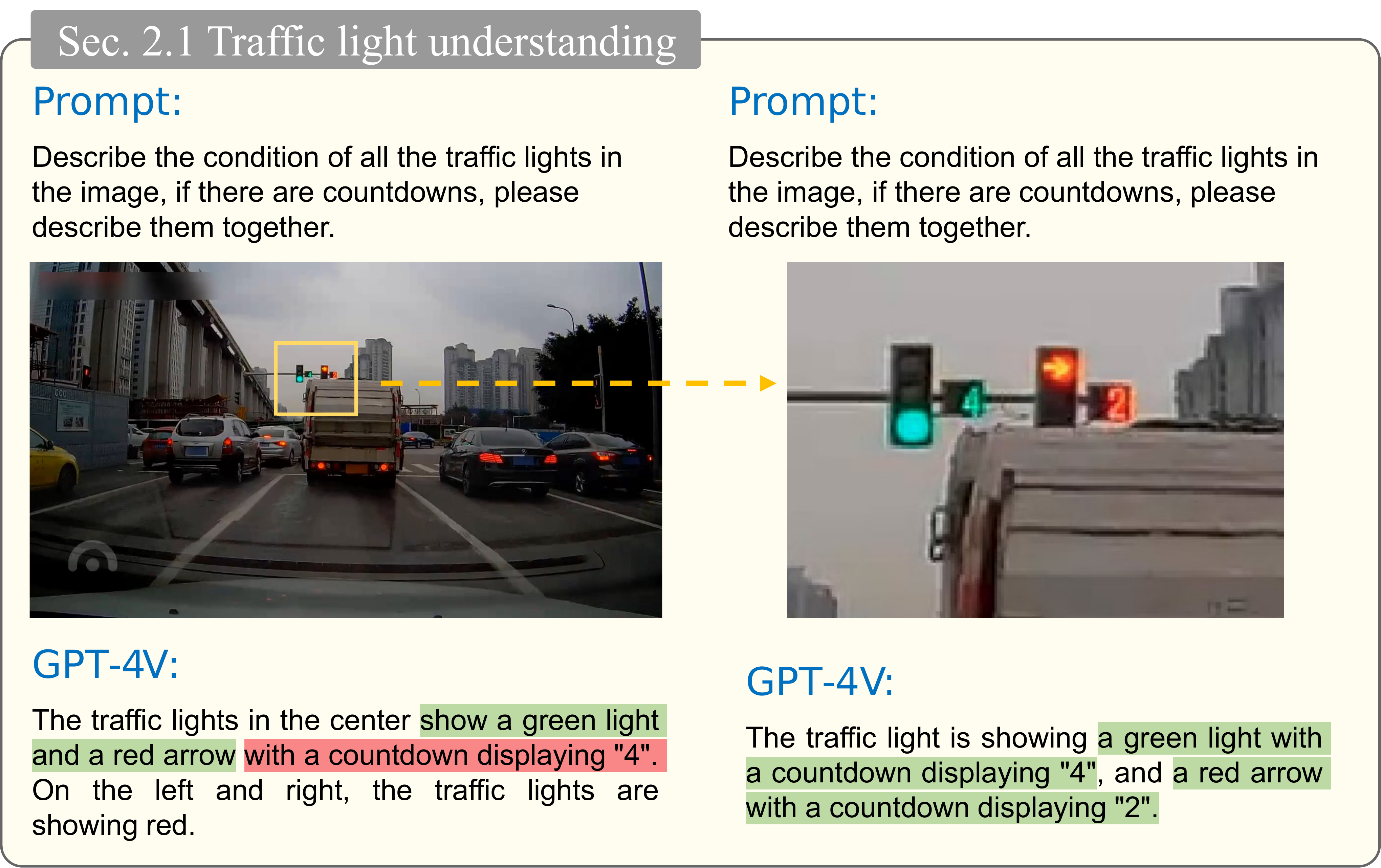}
    \caption[Section~\ref{sec:environmentUnderstanding}: Traffic light understanding, Part II]{Results on the ability to understand the traffic light. \colorbox[RGB]{189,218,165}{Green} highlights the right answer in understanding, \colorbox[RGB]{250,156,154}{Red} highlights the wrong answer in understanding. Check Section~\ref{sec:environmentUnderstanding} for detailed discussions.}
    \label{fig: sec2.1 traffic light understanding 2}
\end{figure}

\textbf{Traffic light understanding.} 
Recognition of traffic lights plays a pivotal role in the functionality of an autonomous driving system. Incorrectly identifying or missing traffic lights not only leads to violations of traffic regulations but also poses a serious risk of traffic accidents. Unfortunately, the performance of \modelname in this test falls short, as evident in Figure~\ref{fig: sec2.1 traffic light understanding 1} and Figure~\ref{fig: sec2.1 traffic light understanding 2}.
In Figure~\ref{fig: sec2.1 traffic light understanding 1}, \modelname demonstrates proficiency in distinguishing between yellow street lights and red traffic lights, particularly during nighttime conditions. However, in Figure~\ref{fig: sec2.1 traffic light understanding 2}, when confronted with a smaller traffic light with a countdown timer in the distant part of the image, \modelname inaccurately identifies the countdown as red and overlooks the genuine 2-second red countdown. The model can provide the correct response only when the traffic light is zoomed in to occupy a significant portion of the image.
Furthermore, \modelname exhibited instances of misrecognition of traffic lights during subsequent tests, which is deemed unacceptable for a mature autonomous driving system.

\begin{figure}[htbp]
    \centering
    \includegraphics[width=\textwidth]{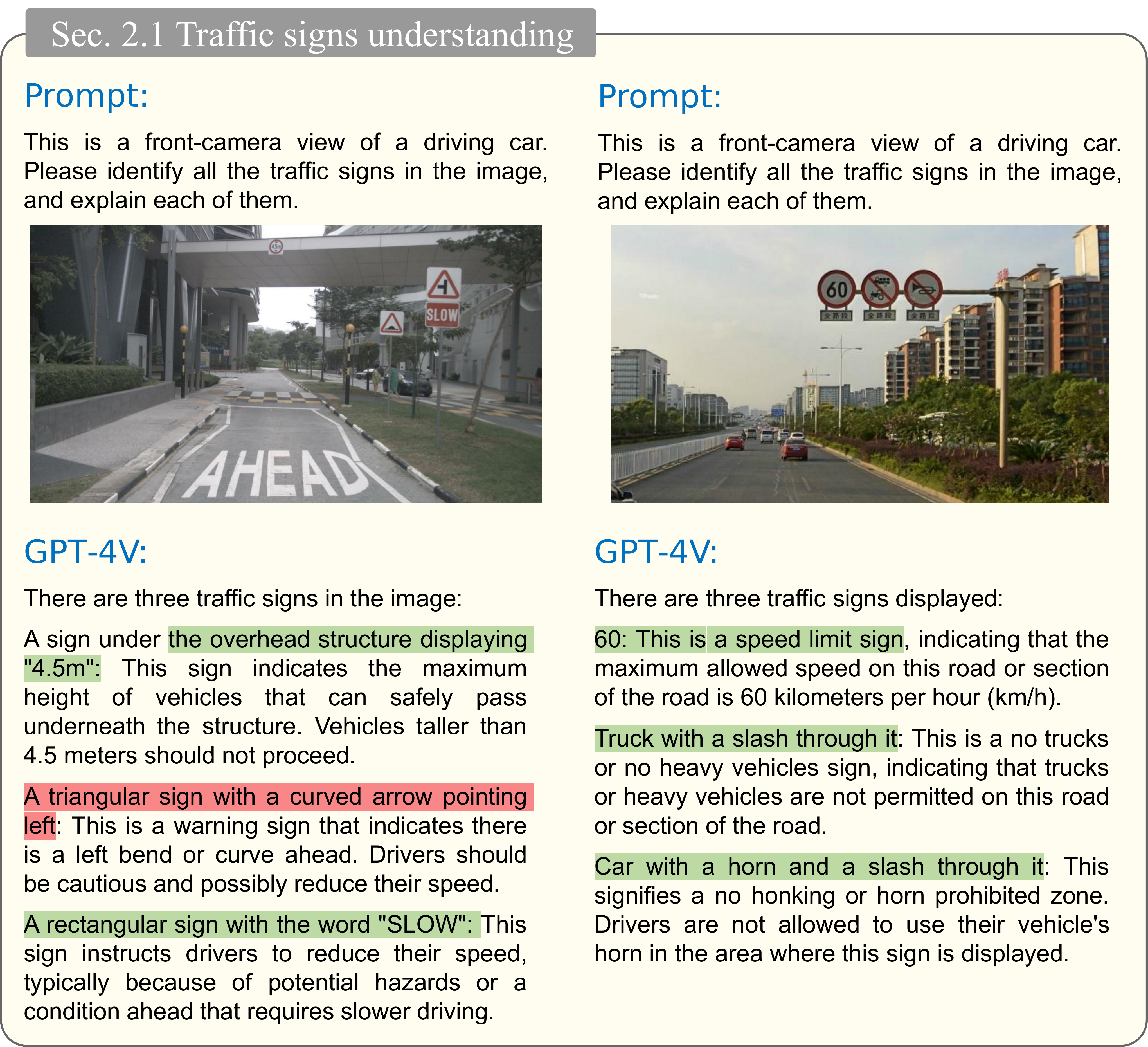}
    \caption[Section~\ref{sec:environmentUnderstanding}: Understanding of traffic signs]{Results on the ability to understand the traffic signs. \colorbox[RGB]{189,218,165}{Green} highlights the right answer in understanding, \colorbox[RGB]{250,156,154}{Red} highlights the wrong answer in understanding. Check Section~\ref{sec:environmentUnderstanding} for detailed discussions.}
    \label{fig: sec2.1 traffic sign understanding}
\end{figure}

\textbf{Traffic signs understanding.}
Traffic signs contain various rules and instructions that drivers need to follow. Autonomous driving systems can understand and comply with these rules by identifying traffic signs, thereby reducing the risk of traffic accidents and improving driving safety. Therefore, we selected representative images from Singapore and China for testing. As can be seen from the left sample in Figure~\ref{fig: sec2.1 traffic sign understanding}, the model can recognize most road signs, including the nearby ``SLOW'' and the distant height limit ``4.5m'', but it misrecognizes the ``Speed Bump'' sign. The three signs from the right sample are all correctly recognized. This indicates that \modelname possesses a commendable ability to recognize traffic signs, yet there remains room for further enhancement.

\clearpage
\subsection{Understanding of Traffic Participants}
\label{sec:trafficParticipantsUnderstanding}

Accurately understanding the status and behavior of traffic participants is the foundation of driving. Existing autonomous driving systems often use a variety of cameras and sensors to perceive traffic participants in order to obtain more comprehensive information about them.
In this section, we assess \modelname's proficiency in comprehending the behavior of traffic participants using various sensor inputs, including 2D images, visualizations of 3D point clouds, and images acquired from V2X devices and autonomous driving simulation software.
The visual data employed here is drawn from nuScenes~\cite{caesar2020nuscenes}, ADD~\cite{wu2023add}, Waymo~\cite{sun2020scalability}, DAIR-V2X~\cite{yu2022dair}, CitySim~\cite{zheng2022citysim} and Carla~\cite{dosovitskiy2017carla} simulation.

\begin{figure*}[!b]
    \centering
    \includegraphics[width=\textwidth]{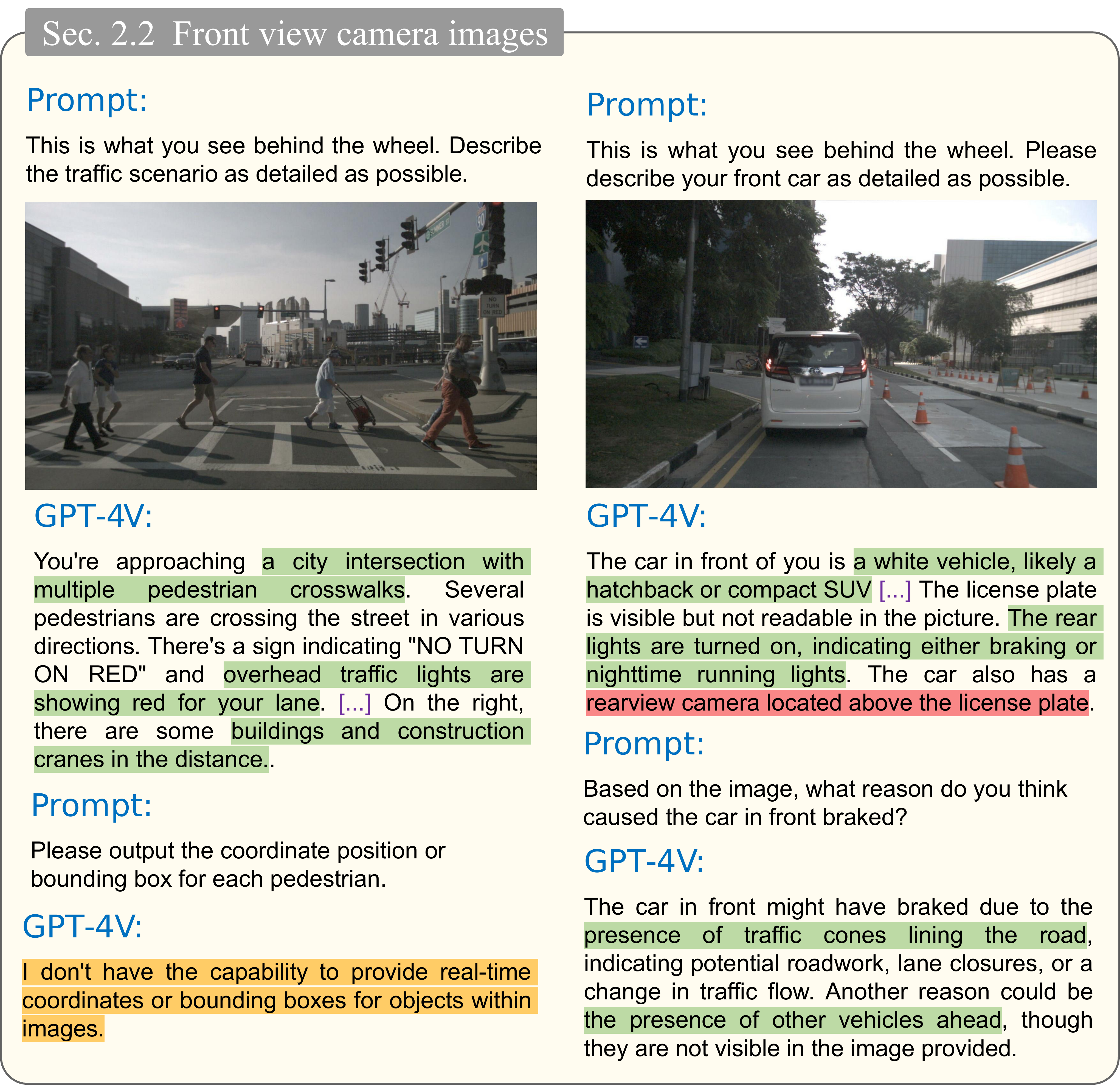}
    \caption[Section~\ref{sec:trafficParticipantsUnderstanding}: Front-view camera images, Part I]{Results describing traffic participants via the front-view camera photo. \colorbox[RGB]{189,218,165}{Green} highlights the right answer in understanding, \colorbox[RGB]{250,156,154}{Red} highlights the wrong answer in understanding, \colorbox[RGB]{255, 204, 102}{Yellow} highlights the incompetence in performing the task. Check Section~\ref{sec:trafficParticipantsUnderstanding} for detailed discussions.
}
    \label{fig:sec2.2 front view 1}
\end{figure*}

\begin{figure*}[htbp]
    \centering
    \includegraphics[width=\textwidth]{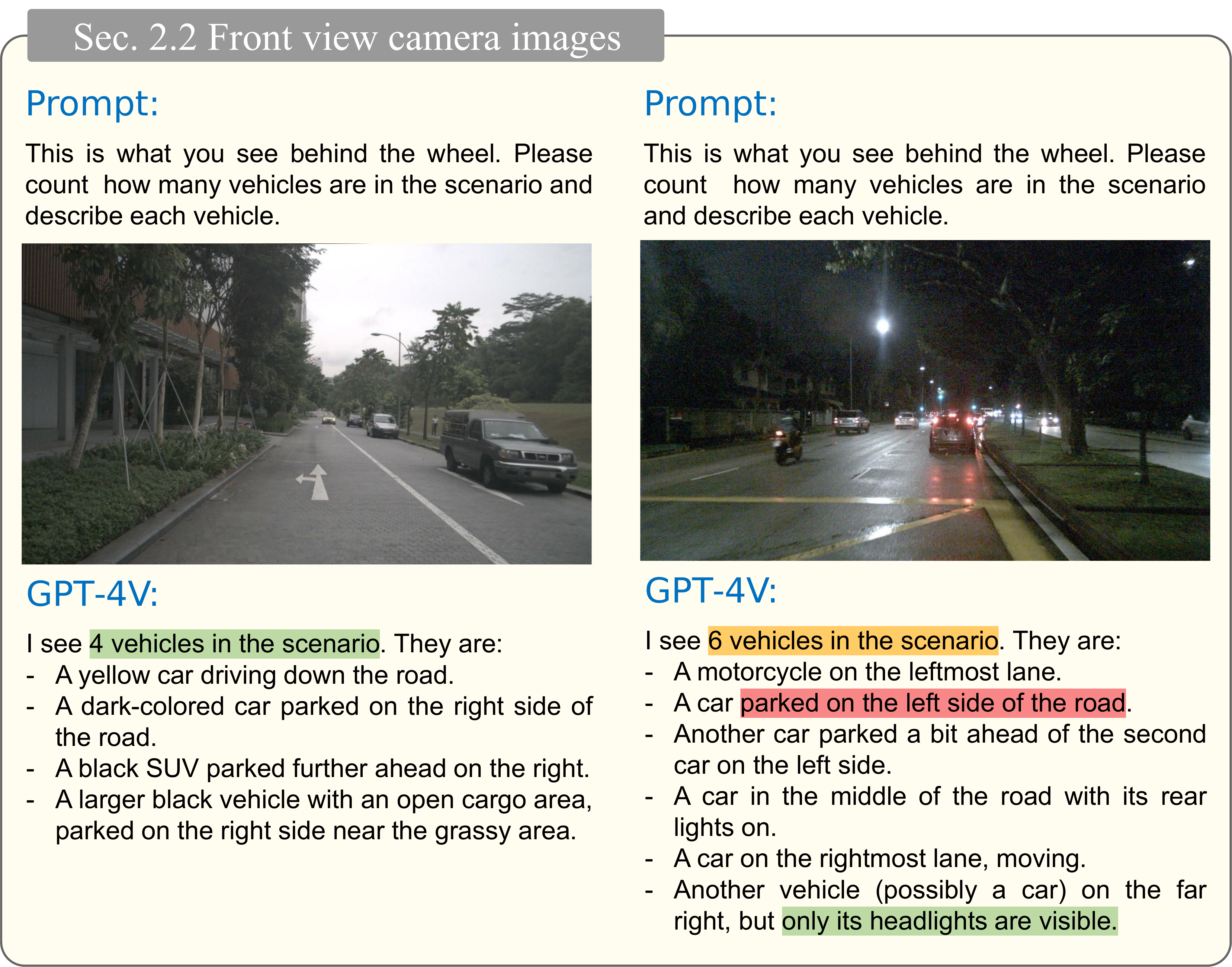}
    \caption[Section~\ref{sec:trafficParticipantsUnderstanding}: Front-view camera images, Part II]{Results describing traffic participants via the front-view camera photo. \colorbox[RGB]{189,218,165}{Green} highlights the right answer in understanding, \colorbox[RGB]{250,156,154}{Red} highlights the wrong answer in understanding, \colorbox[RGB]{255, 204, 102}{Yellow} highlights that the result is correct but the details are wrong. Check Section~\ref{sec:trafficParticipantsUnderstanding} for detailed discussions.
}
    \label{fig:sec2.2 front view 2}
\end{figure*}

\textbf{Front-view camera photos.}
In order to test the model's basic recognition capabilities, including traffic participant recognition and vehicle counting, we input a series of front views of driving scenes and obtained the output results of \modelname. As can be seen from the left side of Figure~\ref{fig:sec2.2 front view 1}, the model can completely and accurately describe the driving scene: it recognizes pedestrians, traffic signs, traffic light status and the surrounding environment. The right side of Figure~\ref{fig:sec2.2 front view 1} shows that the model can identify the vehicle type and its taillights, and can guess its intention to turn on the taillights. However, the model outputs some incorrect statements in irrelevant places, such as thinking that the car in front has a rearview camera.
In Figure~\ref{fig:sec2.2 front view 2}, the counting proficiency of \modelname is put to the test. Utilizing both a daytime and a nighttime snapshot from the vehicle's front view, the model precisely tallies the vehicular presence and their statuses in the daylight capture. In contrast, during nighttime conditions, despite \modelname accurately enumerating the discernible vehicles, its elaborate description of each individual vehicle sometimes falls short of accuracy.

\begin{figure*}[htbp]
    \centering
    \includegraphics[width=\textwidth]{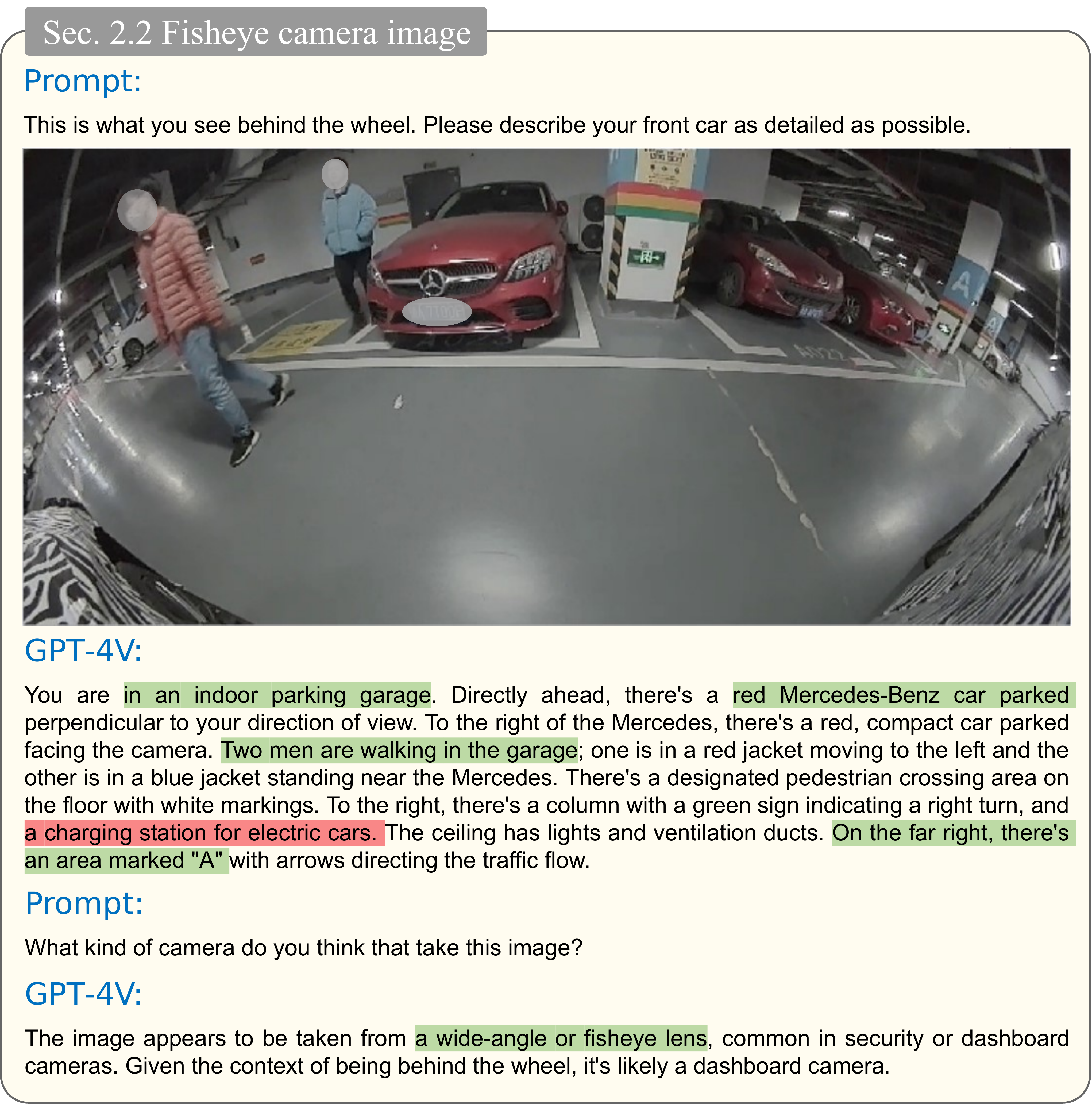}
    \caption[Section~\ref{sec:trafficParticipantsUnderstanding}: Fisheye camera images]{Results describing traffic participants via the photos taken by a fisheye camera. \colorbox[RGB]{189,218,165}{Green} highlights the right answer in understanding, \colorbox[RGB]{250,156,154}{Red} highlights the wrong answer in understanding. Check Section~\ref{sec:trafficParticipantsUnderstanding} for detailed discussions.
}
    \label{fig:sec2.2 fisheye view}
\end{figure*}

\textbf{Fish-eye camera photo.}
The fisheye camera, a prevalent imaging device within autonomous vehicle systems, was also employed to evaluate the perception abilities of \modelname. Results derived from images captured by a fisheye lens are documented in Figure~\ref{fig:sec2.2 fisheye view}. \modelname exhibits an impressive robust tolerance for the distinctive fisheye distortion and shows a commendable understanding of the indoor parking environment. It reliably identifies parked vehicles and the presence of pedestrians in proximity, although there are hallucinations describing a charging station that doesn't exist. Moreover, when queried about the potential apparatus used to take the photo, \modelname accurately discerns it as the work of a fisheye camera.

\begin{figure*}[htbp]
    \centering
    \includegraphics[width=\textwidth]{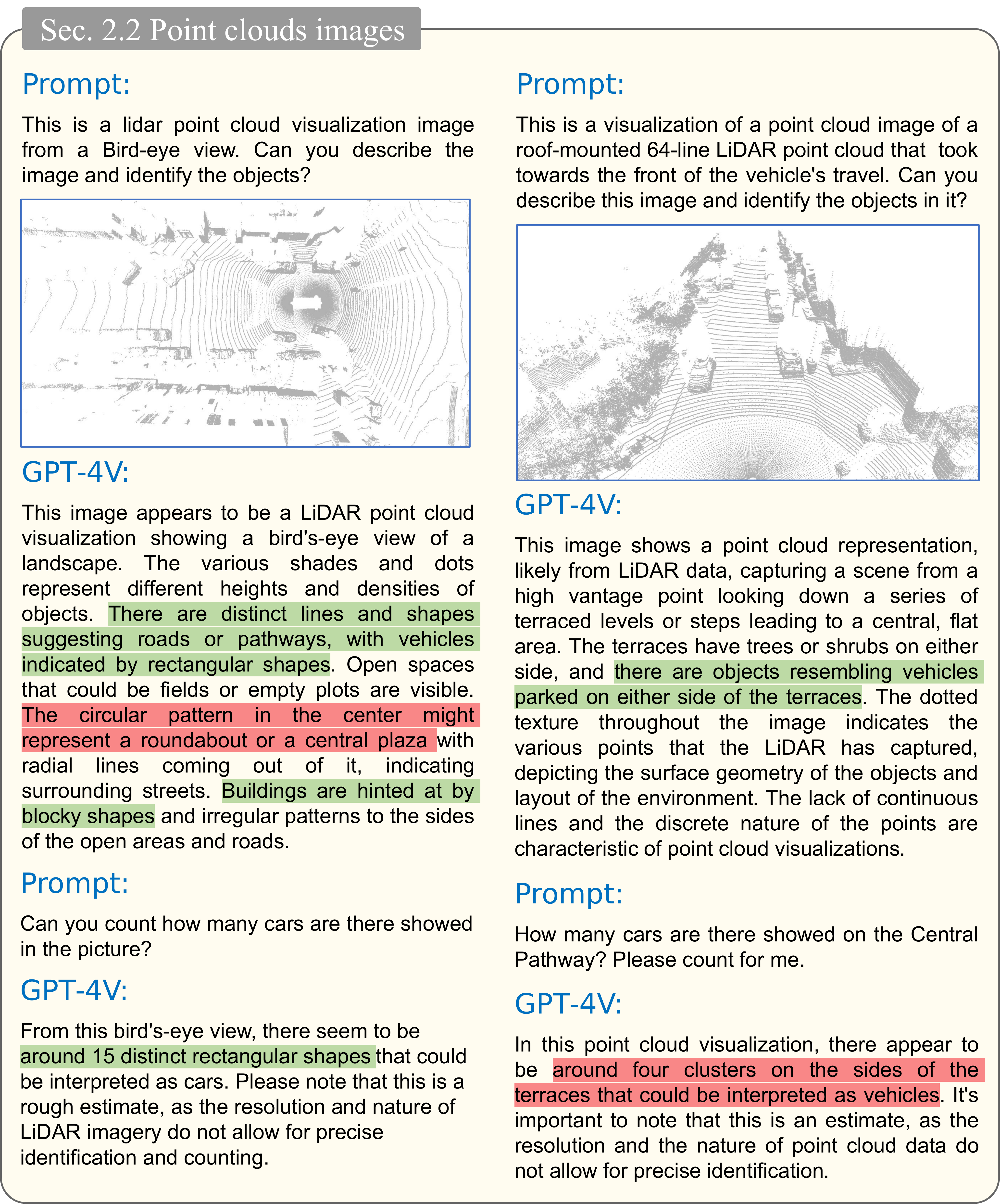}
    \caption[Section~\ref{sec:trafficParticipantsUnderstanding}: Point cloud lidar images]{Results describing traffic participants via visualization image from the point cloud taken by lidar. \colorbox[RGB]{189,218,165}{Green} highlights the right answer in understanding, \colorbox[RGB]{250,156,154}{Red} highlights the wrong answer in understanding. Check Section~\ref{sec:trafficParticipantsUnderstanding} for detailed discussions.
}
    \label{fig:sec2.2 point cloud view}
\end{figure*}

\textbf{Point cloud visualization images.}
Out of curiosity, we captured two screenshots of a 64-line LiDAR point cloud, one from the bird's-eye view and the other from the front view. Although compressing the point cloud on a two-dimensional image will inevitably lose the three-dimensional geometric information, several distinctive features can still be discerned and classified. The test is shown in Figure~\ref{fig:sec2.2 point cloud view}. Subsequently, we feed these two images into \modelname, and to our surprise, it exhibits the capability to recognize certain road and building patterns within them. Since the model has rarely seen this type of data before, it inevitably assumed that the circular pattern in the bird's-eye view represented a roundabout or a central plaza. Furthermore, when tasked with identifying vehicles, the model is largely successful in estimating the number of vehicles in the scene. We also observe counting errors in the front view, which are caused by the outlines of some vehicles being incomplete and difficult to discern. Through this test, the powerful ability of the model to process unconventional data is demonstrated.

\begin{figure*}[htbp]
    \centering
    \includegraphics[width=\textwidth]{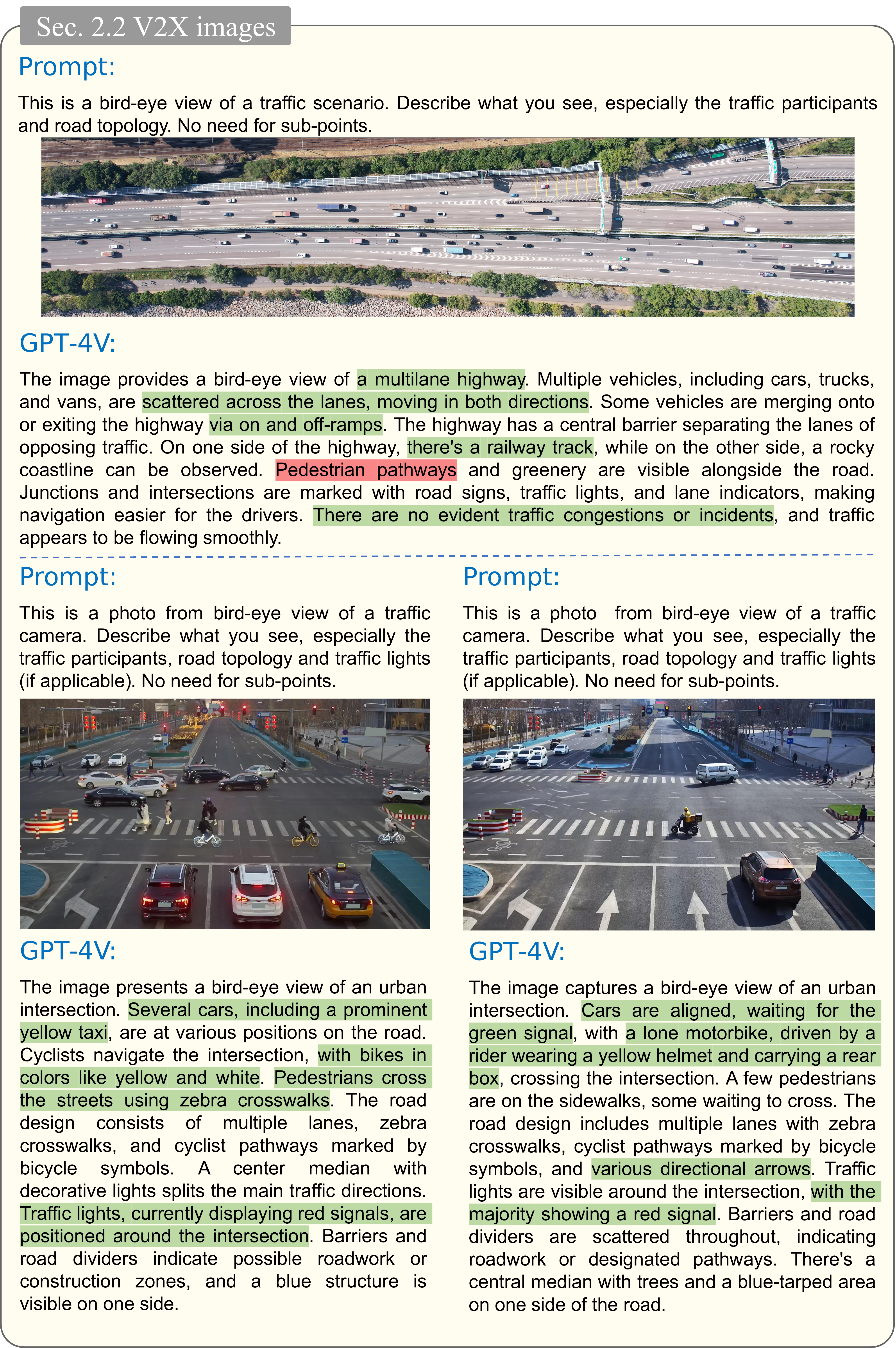}
    \caption[Section~\ref{sec:trafficParticipantsUnderstanding}: V2X equipment images]{Results describing traffic participants via the V2X equipment's photos. \colorbox[RGB]{189,218,165}{Green} highlights the right answer in understanding, \colorbox[RGB]{250,156,154}{Red} highlights the wrong answer in understanding. Check Section~\ref{sec:trafficParticipantsUnderstanding} for detailed discussions.
}
    \label{fig:sec2.2 v2x view}
\end{figure*}

\textbf{V2X devices photos.} 
V2X, which stands for Vehicle-to-Everything, encompasses a range of technologies that enable vehicles to communicate with not only each other but also with infrastructure and various other entities. V2X cameras play a pivotal role in capturing and processing visual information as part of this interconnected ecosystem.
In Figure~\ref{fig:sec2.2 v2x view}, we present the responses generated by \modelname for a drone-view photograph and two intersection camera images. \modelname exhibits commendable performance in all three instances.
In the drone view, \modelname accurately identifies the freeway in both directions and recognizes the on-ramp situated on the right side of the photo. 
And in the intersection V2X device view, the response identifies a mixed traffic flow containing cars, cyclists, and pedestrians in the image, as well as accurate traffic light recognition.

\begin{figure*}[htbp]
    \centering
    \includegraphics[width=\textwidth]{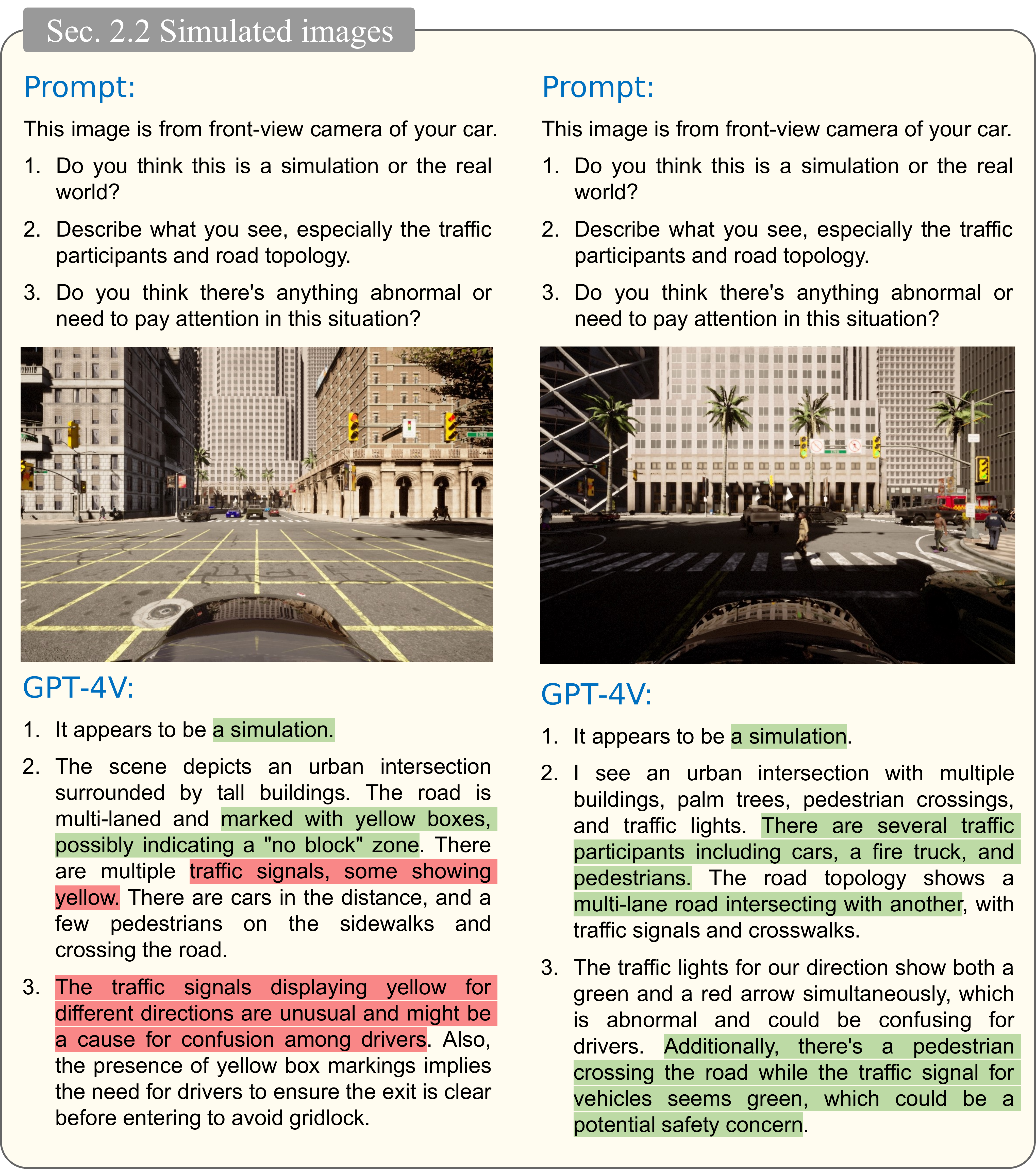}
    \caption[Section~\ref{sec:trafficParticipantsUnderstanding}: Simulated images]{Results describing traffic participants via the images taken in the CARLA simulator. \colorbox[RGB]{189,218,165}{Green} highlights the right answer in understanding, \colorbox[RGB]{250,156,154}{Red} highlights the wrong answer in understanding. Check Section~\ref{sec:trafficParticipantsUnderstanding} for detailed discussions.
}
    \label{fig:sec2.2 simulated view}
\end{figure*}

\textbf{Images taken in CARLA simulator.}
In the realm of autonomous driving research and development, simulators like CARLA serve as invaluable tools, providing a controlled and virtual environment where algorithms can be tested, trained, and refined before their deployment on real-world roads~\cite{dosovitskiy2017carla}.
We captured a series of images within CARLA's map of the Town 10, using the ego car as the primary viewpoint. Subsequently, we posed several questions based on these images, and the outcomes are showcased in Figure~\ref{fig:sec2.2 simulated view}.
\modelname not only identifies these images as originating from simulation software but also demonstrates a high level of awareness regarding the virtual vehicles and pedestrians within them. Furthermore, in rare instances where simulated pedestrians run red lights, \modelname appropriately acknowledges this scenario in its responses. However, it's worth noting that the model still struggles with recognizing traffic lights in simulation, like misidentifying red lights as yellow.

%% file: sections/reasoning.tex
\clearpage
\section{Advanced Capability of Reasoning}
\label{sec:reasoning}

Reasoning is another important trait for proper driving behavior. Given the dynamic and unpredictable nature of traffic environments, drivers often encounter a range of unexpected events. It is imperative for proficient drivers to make accurate judgments and decisions, drawing on their experience and common sense when faced with such unforeseen circumstances. In this section, we have carried out a series of tests to evaluate \modelname's responses to unexpected events and its proficiency in navigating dynamic environments.

\subsection{Corner Cases}
\label{sec:cornerCases}

In the ongoing research for autonomous driving, the common approach to enhancing the system's ability to handle unexpected events is through the continuous collection of data. However, given the dynamic, continuous, and random nature of driving environments, the data collected can only approximate its boundaries but never fully encapsulate it, that is, unexpected events are inevitable. Human drivers, equipped with common sense, are often able to improvise and navigate safely through these unforeseen circumstances. This highlights the importance of incorporating not just data-driven methods, but also the principles of reasoning and common sense into autonomous driving systems. 
The visual data employed here is drawn from CODA~\cite{li2022coda} and the internet.

In this section, we have carefully curated a set of perceptual corner cases to assess the model's capacity for common-sense reasoning. These examples deliberately include objects that fall out of the typical distribution, often posing challenges for conventional perception systems and creating difficulties in decision-making planning. Now, let's see how \modelname fares in addressing these cases.

\begin{figure*}[htbp]
    \centering
    \includegraphics[width=\textwidth]{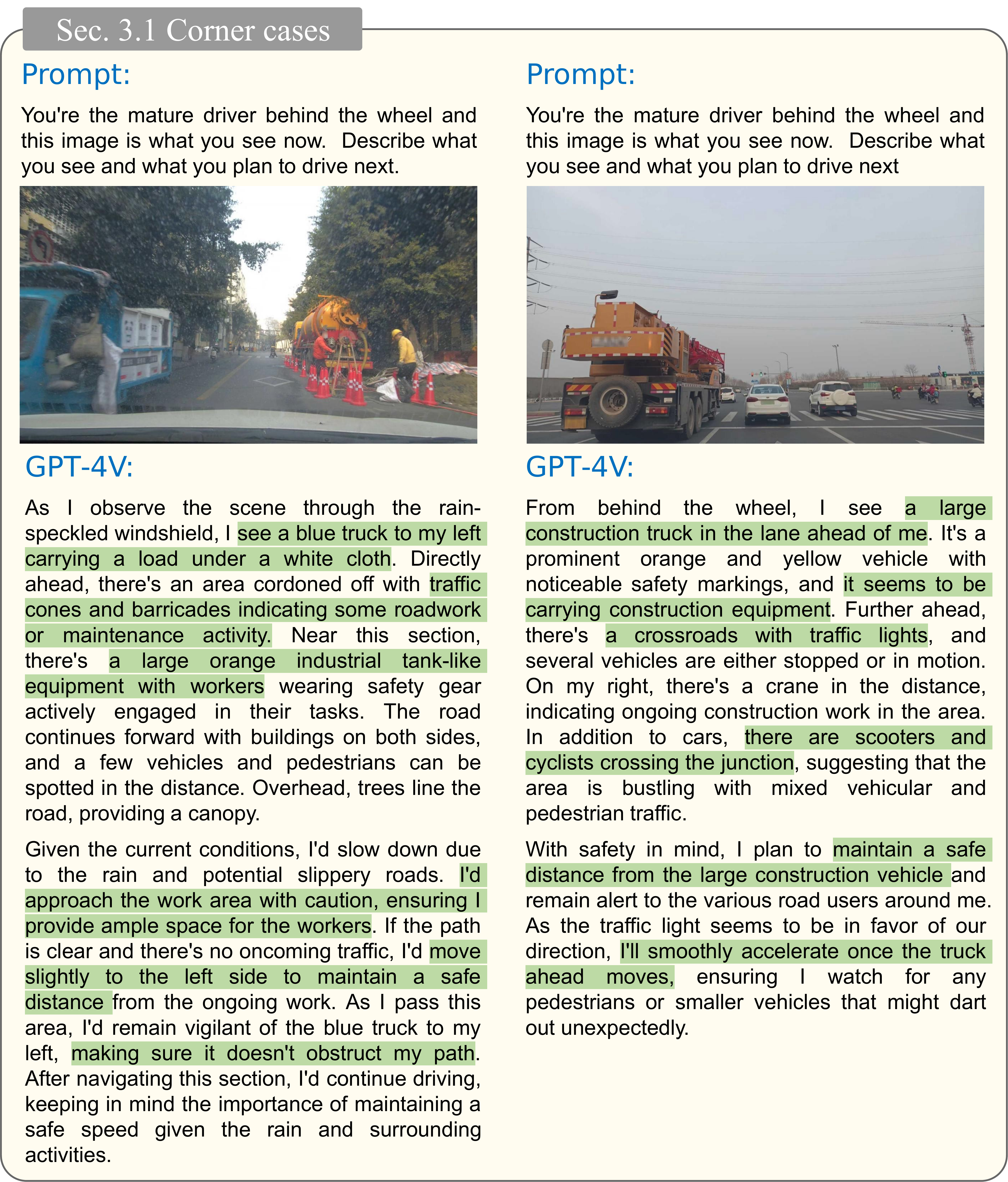}
    \caption[Section~\ref{sec:cornerCases}: Corner cases, Part I]{Illustration of \modelname's ability to make correct decisions in corner cases. \colorbox[RGB]{189,218,165}{Green} highlights the right answer in understanding. Check Section~\ref{sec:cornerCases} for detailed discussions.}
    \label{fig: sec3.1 corner cases1}
\end{figure*}

On the left side of Figure~\ref{fig: sec3.1 corner cases1}, \modelname can clearly describe the appearance of the vehicles that are not commonly seen, the traffic cone on the ground, and the staff beside the vehicle. After identifying these conditions, the model realizes that the ego car can move slightly to the left, maintain a safe distance from the work area on the right, and drive cautiously. 
In the right example, \modelname adeptly identifies a complex traffic scenario, encompassing an orange construction vehicle, sidewalks, traffic lights, and cyclists. When queried about its driving strategy, it articulates an intention to maintain a safe distance from the construction vehicle and, upon its passage, execute a smooth acceleration while conscientiously observing pedestrian presence.

\begin{figure*}[htbp]
    \centering
    \includegraphics[width=\textwidth]{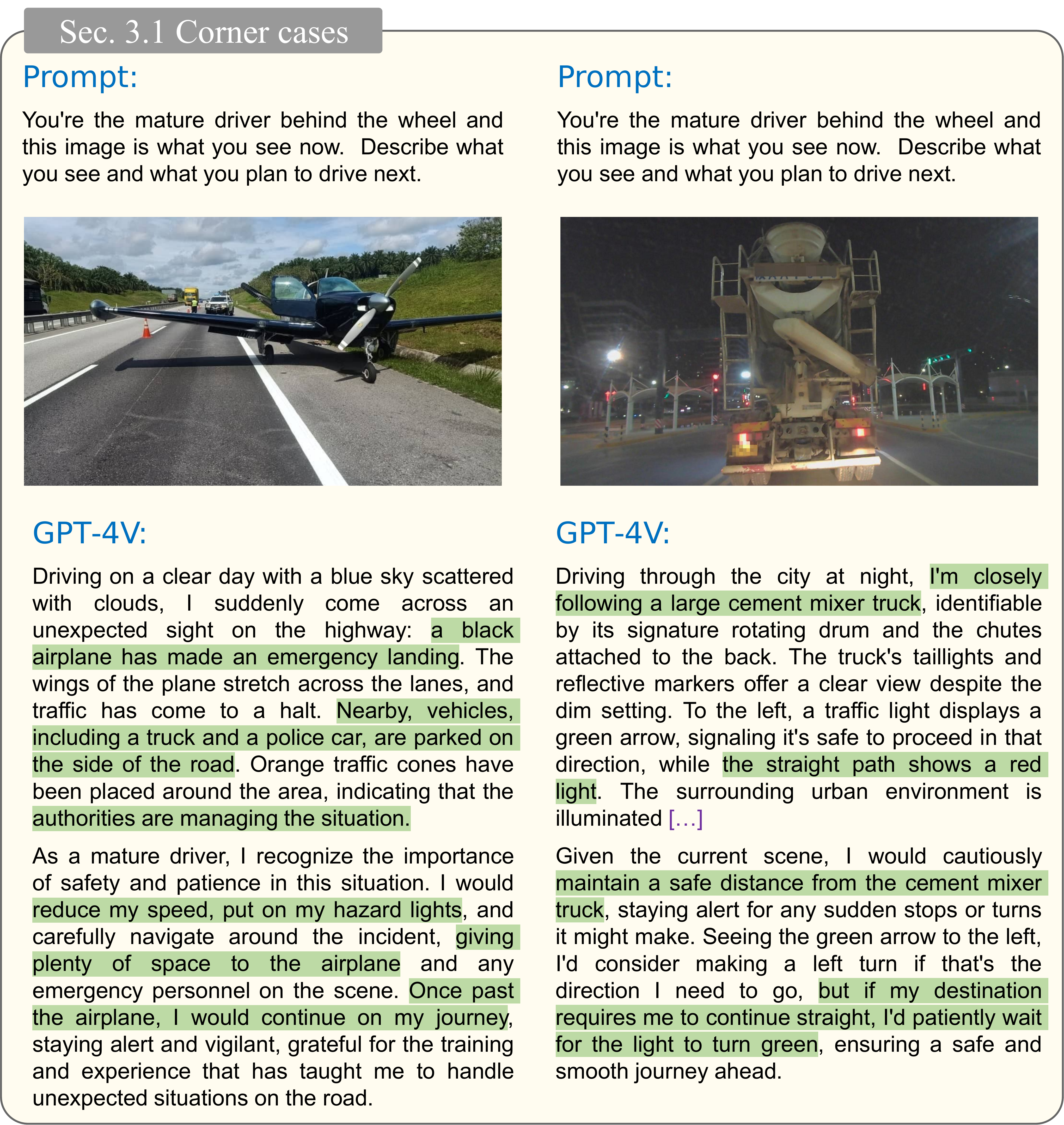}
    \caption[Section~\ref{sec:cornerCases}: Corner cases, Part II]{Illustration of \modelname's ability to make correct decisions in corner cases. \colorbox[RGB]{189,218,165}{Green} highlights the right answer in understanding. Check Section~\ref{sec:cornerCases} for detailed discussions.}
    \label{fig: sec3.1 corner cases2}
\end{figure*}

On the left side of Figure~\ref{fig: sec3.1 corner cases2}, \modelname can accurately identify that an airplane has made an emergency landing on the road and authorities are handling the situation nearby. For traditional perception algorithms, it is difficult to recognize without specific training. Under these conditions, the model knows that it should slow down and turn on its hazard lights while awaiting clearance to pass the aircraft before resuming regular driving.
In the right example, \modelname accurately identifies the cement mixer truck and the red traffic light ahead. It discerns the importance of maintaining a safe distance from the truck in the lead until the red light transitions to green, at which point it proceeds with its journey.

\begin{figure*}[htbp]
    \centering
    \includegraphics[width=\textwidth]{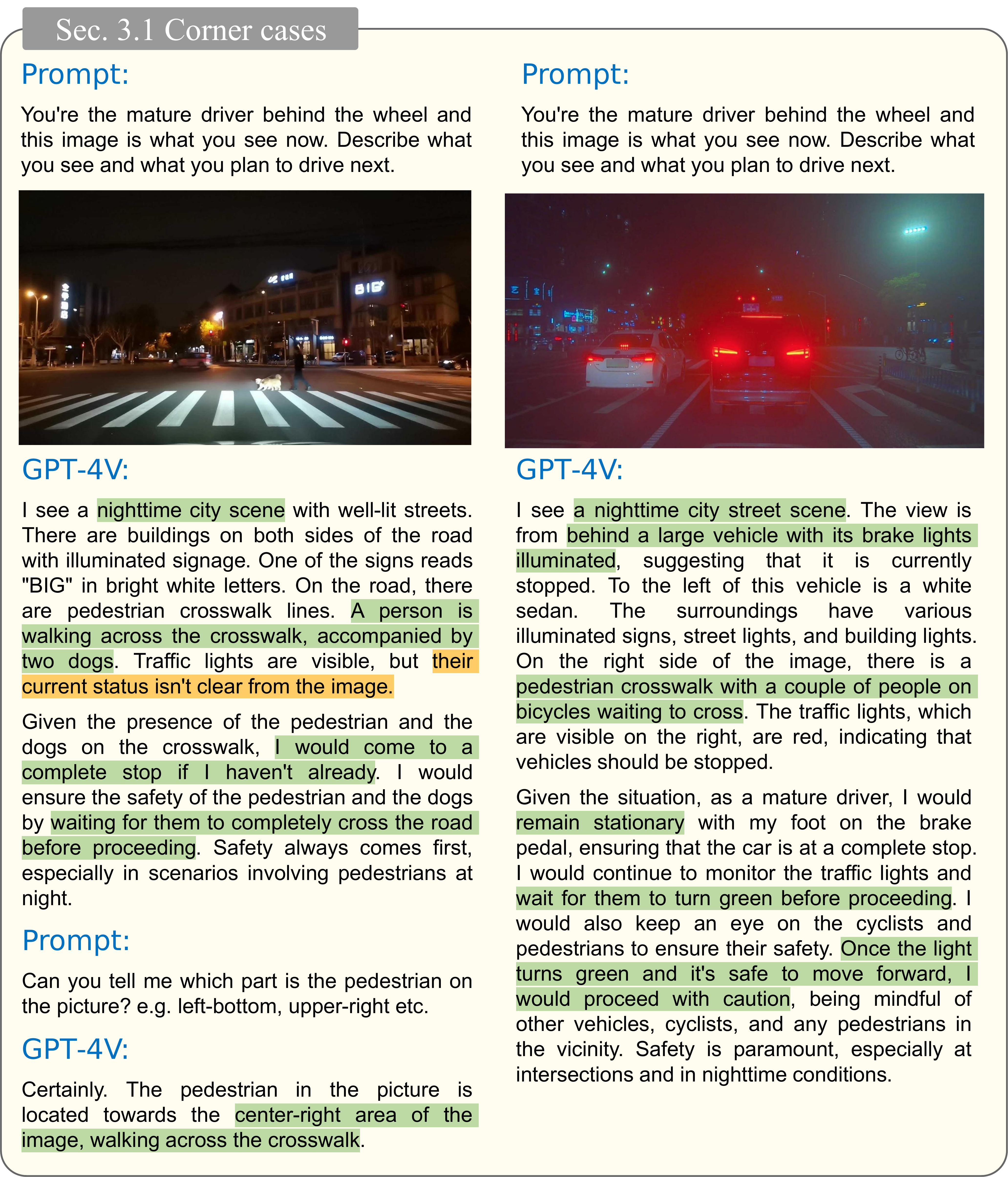}
    \caption[Section~\ref{sec:cornerCases}: Corner cases, Part III]{Illustration of \modelname's ability to make correct decisions in corner cases. \colorbox[RGB]{189,218,165}{Green} highlights the right answer in understanding, \colorbox[RGB]{255, 204, 102}{Yellow} highlights the incompetence in performing the task. Check Section~\ref{sec:cornerCases} for detailed discussions.}
    \label{fig: sec3.1 corner cases3}
\end{figure*}

On the left side of Figure~\ref{fig: sec3.1 corner cases3}, \modelname describes a scene where a pedestrian accompanied by two dogs is crossing the crosswalk, positioned in the center-right area of the image. Remarkably, the model accurately counts the number of dogs present. The model concludes that the vehicle should patiently wait for the pedestrian and dogs to clear the way before resuming its journey, though it remains unable to ascertain the status of the traffic light. It's worth mentioning that this isn't the first occurrence of such an incident for \modelname.
The image on the right depicts a nocturnal traffic scenario, which is well recognized by \modelname. In this instance, the model astutely discerned the illuminated brake lights of the vehicle ahead and noted the presence of pedestrians and cyclists, patiently waiting to cross the road. It intelligently deduced that it should remain stationary until the traffic signal turns green, exercising due caution when resuming its journey.

\clearpage
\subsection{Multi-view Images}
\label{sec:multiViewImages}

By utilizing multi-view cameras, \modelname captures a comprehensive view of the driving environment. Precisely interpreting the spatial relationships between these cameras and the overlapping areas within the images is crucial for the model to effectively harness the potential of the multi-view camera system. In this section, we evaluate \modelname's competence in handling multi-view images. All data in this section comes from the nuScenes~\cite{caesar2020nuscenes} dataset.

\begin{figure*}[htbp]
    \centering
    \includegraphics[width=\textwidth]{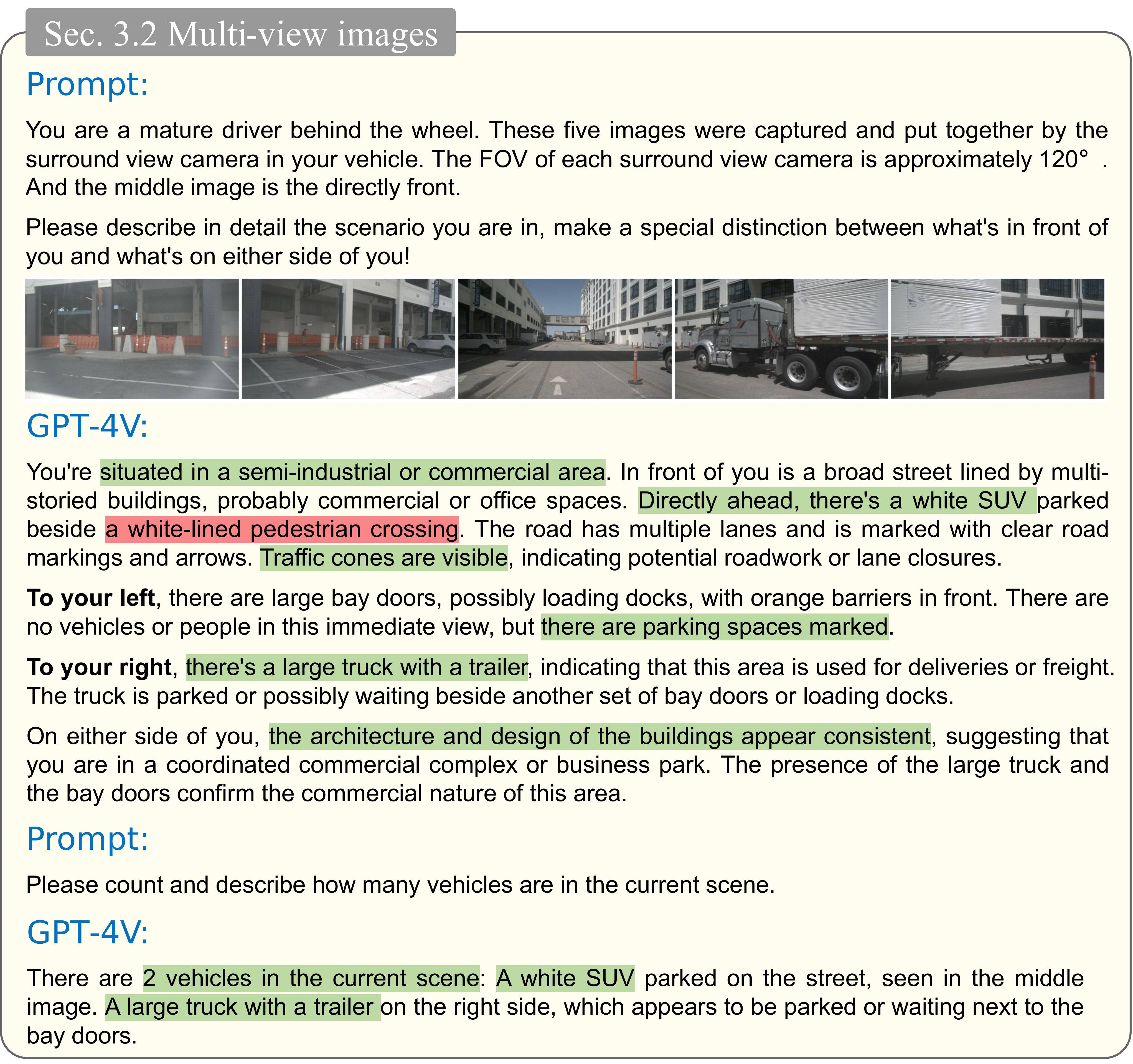}
    \caption[Section~\ref{sec:multiViewImages}: Multi-view images, Part I]{Illustration of \modelname's ability to infer the relationship between multi-view images. \colorbox[RGB]{189,218,165}{Green} highlights the right answer in understanding. Check Section~\ref{sec:multiViewImages} for detailed discussions.}
    \label{fig: sec3.2 multi-view images1}
\end{figure*}

In Figure~\ref{fig: sec3.2 multi-view images1}, we select a set of surrounding images and input them to the model in the correct sequence. The model adeptly recognizes various elements within the scene, such as buildings, vehicles, barriers, and parking. It can even deduce from the overlapping information that there are two cars in the scene, with one white SUV positioned to the front and a truck with a trailer to the right. While the model's performance is generally impressive, an insignificant error is the misidentification of the pedestrian crossing.

\begin{figure*}[htbp]
    \centering
    \includegraphics[width=\textwidth]{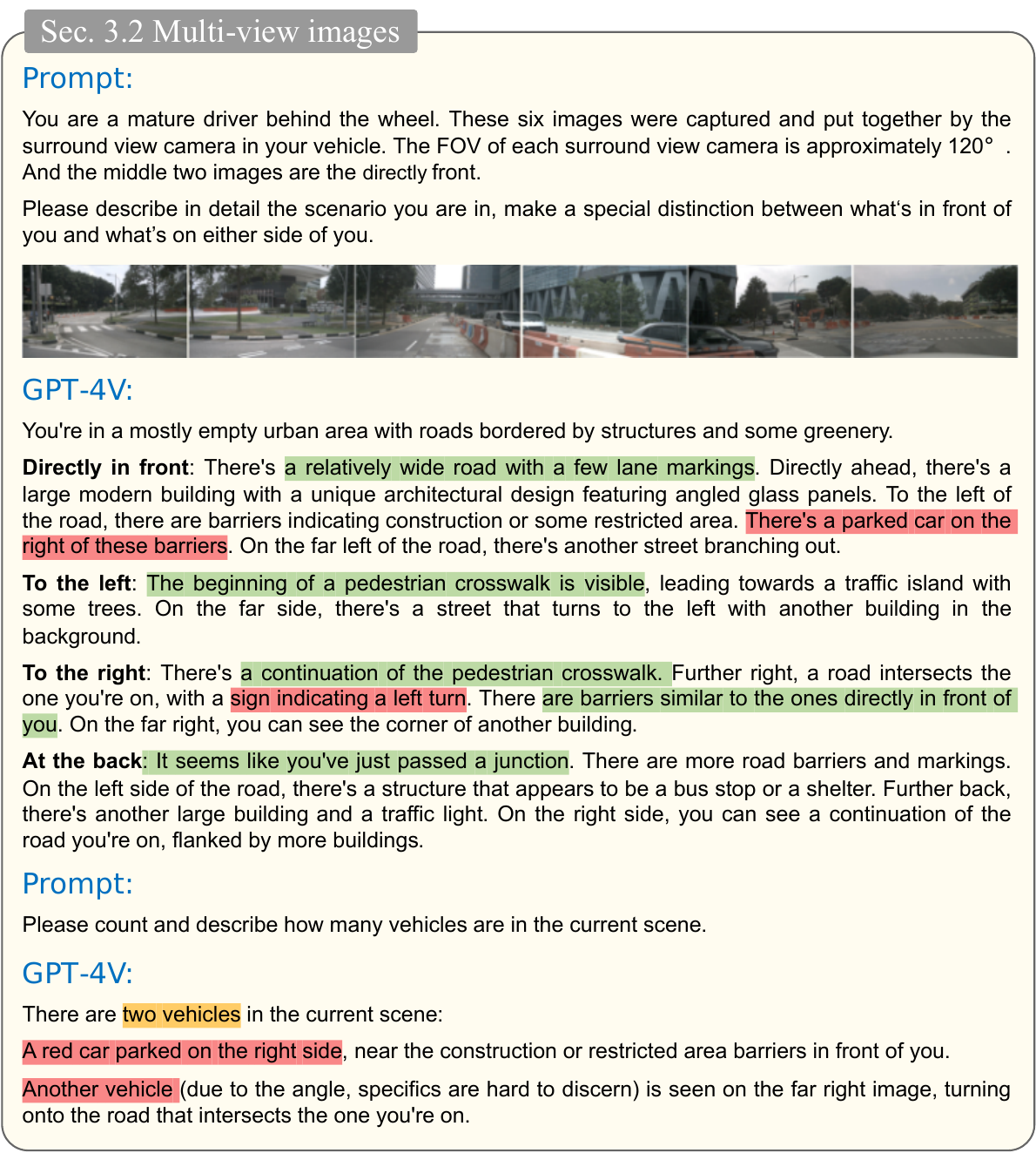}
    \caption[Section~\ref{sec:multiViewImages}: Multi-view images, Part II]{Illustration of \modelname's ability to infer the relationship between multi-view images. \colorbox[RGB]{189,218,165}{Green} highlights the right answer in understanding, \colorbox[RGB]{250,156,154}{Red} highlights the wrong answer in understanding, \colorbox[RGB]{255, 204, 102}{Yellow} highlights the incompetence in performing the task. Check Section~\ref{sec:multiViewImages} for detailed discussions.}
    \label{fig: sec3.2 multi-view images2}
\end{figure*}

In another experiment shown in Figure~\ref{fig: sec3.2 multi-view images2}, we similarly employ a set of combined surrounding images for test. While the model managed to provide a mostly accurate description of the scene, it also exhibited several recognition errors, particularly in regard to the number and shapes of the vehicles present. Notably, the model produces some confusing illusions, such as thinking there is a left-turn sign on the picture. We hypothesize that these issues may arise from the model's limited spatial reasoning capabilities.

\begin{figure*}[htbp]
    \centering
    \includegraphics[width=\textwidth]{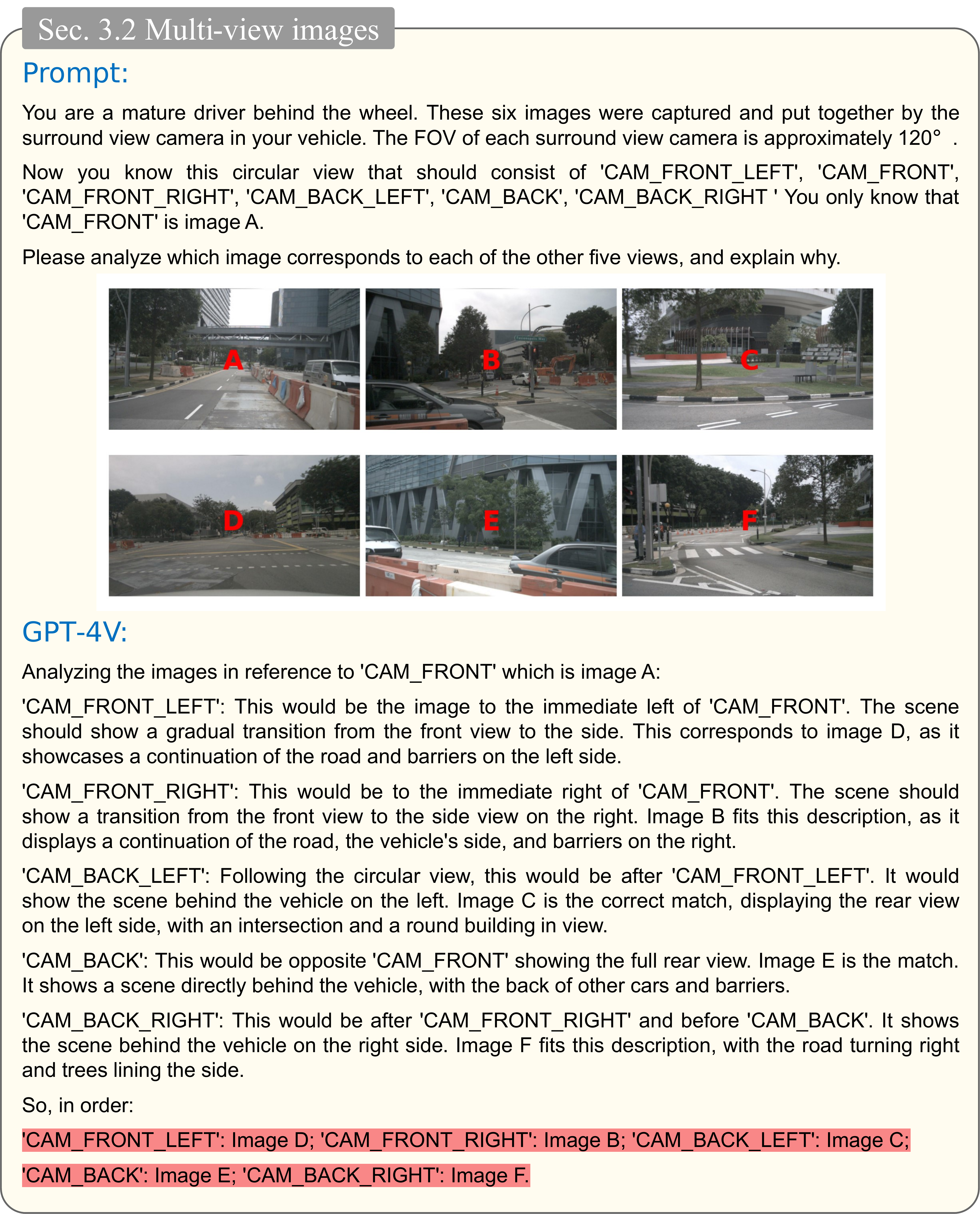}
    \caption[Section~\ref{sec:multiViewImages}: Multi-view images, Part III]{Illustration of \modelname's ability to infer the relationship between multi-view images. \colorbox[RGB]{250,156,154}{Red} highlights the wrong answer in understanding. Check Section~\ref{sec:multiViewImages} for detailed discussions.}
    \label{fig: sec3.2 multi-view images3}
\end{figure*}

Finally, in Figure~\ref{fig: sec3.2 multi-view images3}, we present the correct front view and try to let \modelname identify and sort the out-of-order surrounding images. Although the model performs a lot of analysis and reasoning that seems to make sense, it still ends up outputting all wrong answers. It becomes evident that the model encounters challenges in establishing connections between adjacent images. We acknowledge the intricacies associated with this particular task.

\clearpage
\subsection{Temporal Sequences}
\label{sec:temporalSequences}

In this section, we assess the capability of \modelname in understanding temporal images. Our methodology involves the utilization of multiple sequences from first-person driving videos. From each video segment, we extract four keyframes, label them with sequential numbers, and combine them into a single image for input. Subsequently, we task \modelname with describing what events occurred during this time period, as well as the actions taken by ego vehicle and the reasons behind them. The examples are sourced from nuScenes~\cite{caesar2020nuscenes}, D$^2$-city and Carla~\cite{dosovitskiy2017carla} simulation.

\begin{figure*}[htbp]
    \centering
    \includegraphics[width=\textwidth]{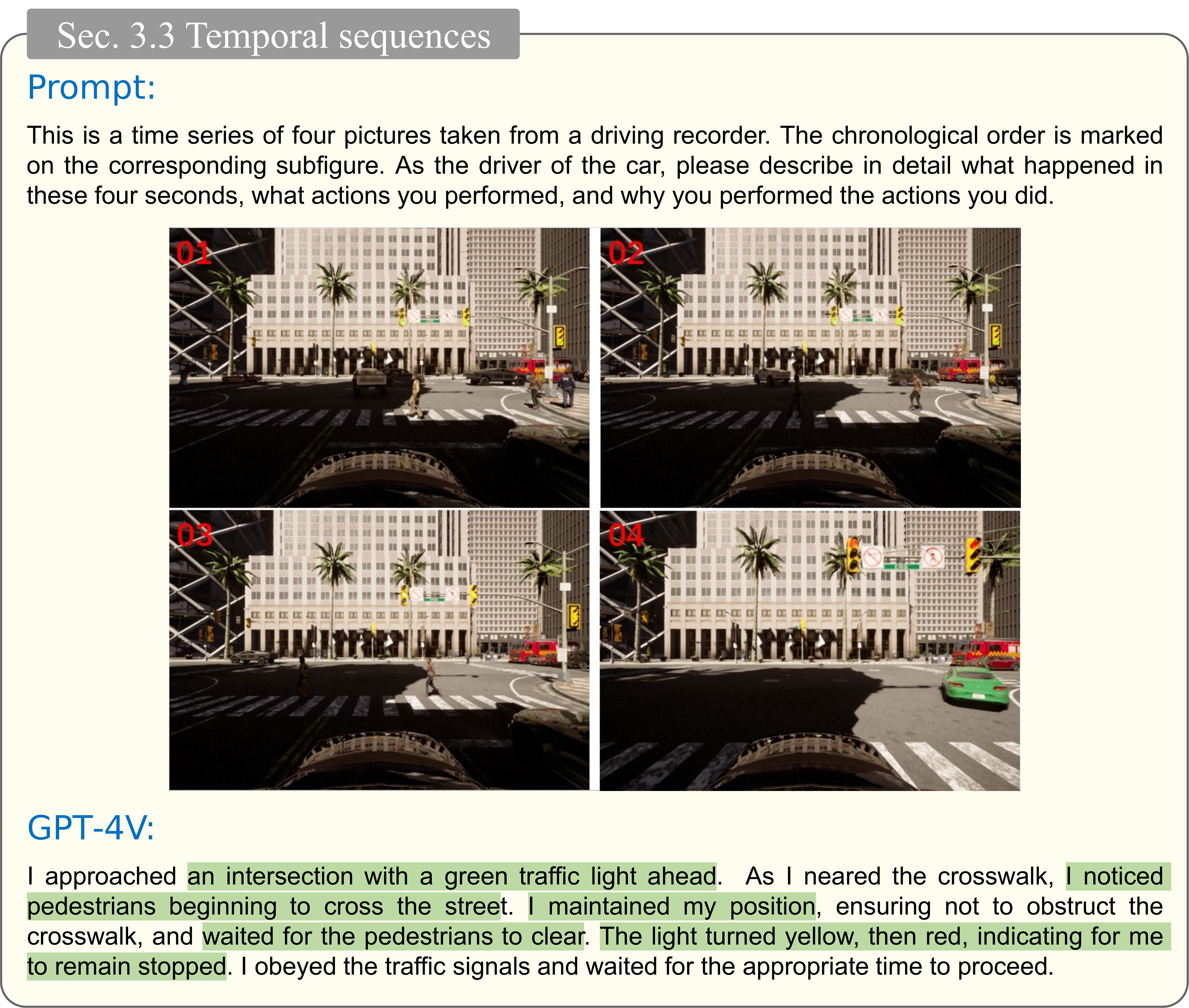}
    \caption[Section~\ref{sec:temporalSequences}: Temporal sequences, Part I]{Illustration of \modelname's ability to understand temporal sequential images.  \colorbox[RGB]{189,218,165}{Green} highlights the right answer in understanding. Check Section~\ref{sec:temporalSequences} for detailed discussions.}
    \label{fig: sec3.3 temporal sequences1}
\end{figure*}

Figure~\ref{fig: sec3.3 temporal sequences1} shows a video captured in CARLA's map of Town 10 where the \modelname clearly explains the action of the ego car stopping at a crosswalk because of a pedestrian crossing the road,  just before the traffic signal turning red.

\begin{figure*}[htbp]
    \centering
    \includegraphics[width=\textwidth]{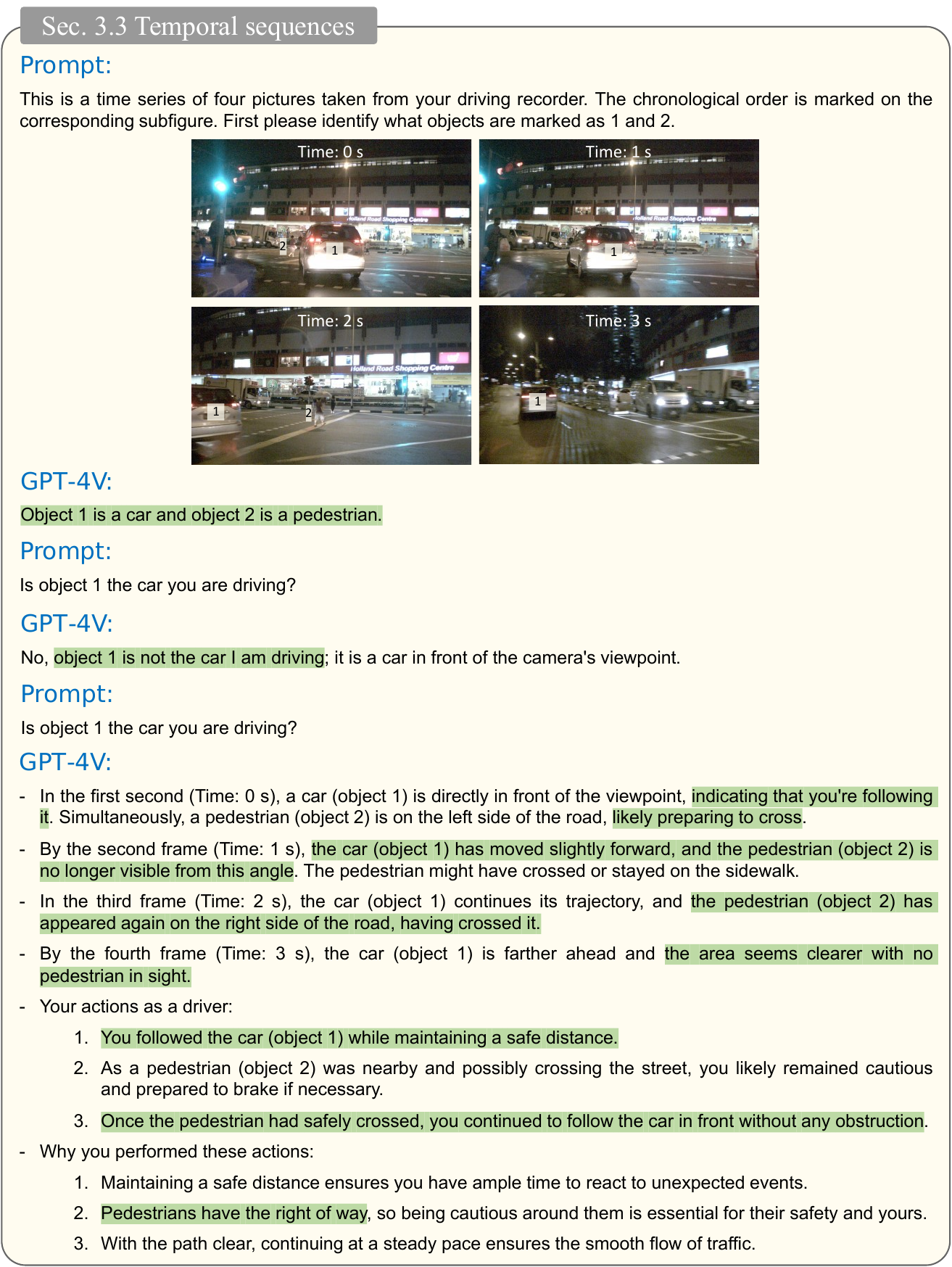}
    \caption[Section~\ref{sec:temporalSequences}: Temporal sequences, Part II]{Illustration of \modelname's ability to understand temporal sequential images.  \colorbox[RGB]{189,218,165}{Green} highlights the right answer in understanding. Check Section~\ref{sec:temporalSequences} for detailed discussions.}
    \label{fig: sec3.3 temporal sequences2}
\end{figure*}

Figure~\ref{fig: sec3.3 temporal sequences2} showcases a video segment extracted from the NuScene dataset~\cite{caesar2020nuscenes}. During the process of capturing keyframes, we assigned the labels ``1'' and ``2'' to the leading SUV and a pedestrian, respectively. \modelname not only responds accurately to inquiries regarding the objects represented by these labels but also offers a comprehensive explanation of the interaction observed between the front SUV and the pedestrian. This interaction entails the pedestrian crossing the street while the white SUV yields the right-of-way.

\begin{figure*}[htbp]
    \centering
    \includegraphics[width=\textwidth]{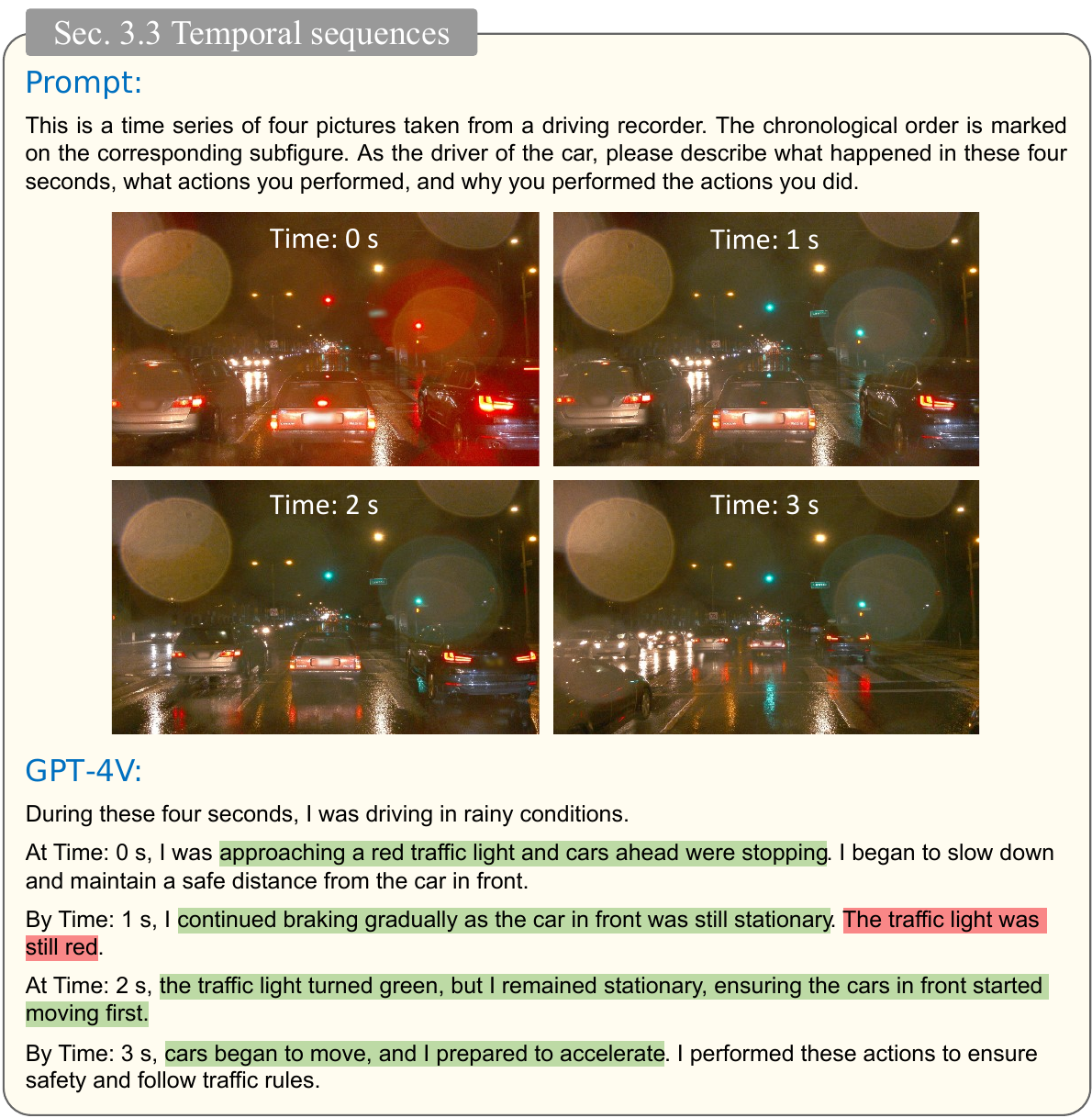}
    \caption[Section~\ref{sec:temporalSequences}: Temporal sequences, Part IV]{Illustration of \modelname's ability to understand temporal sequential images. \colorbox[RGB]{189,218,165}{Green} highlights the right answer in understanding, \colorbox[RGB]{250,156,154}{Red} highlights the wrong answer in understanding. Check Section~\ref{sec:temporalSequences} for detailed discussions.}
    \label{fig: sec3.3 temporal sequences4}
\end{figure*}

The video sequence depicted in Figure~\ref{fig: sec3.3 temporal sequences4} exhibits a noticeable camera flare caused by rain. Nevertheless, \modelname adeptly identifies the intersection signals and the front taillights. It also infers the semantic information about the start of the front car after the red light turns green.

\begin{figure*}[htbp]
    \centering
    \includegraphics[width=\textwidth]{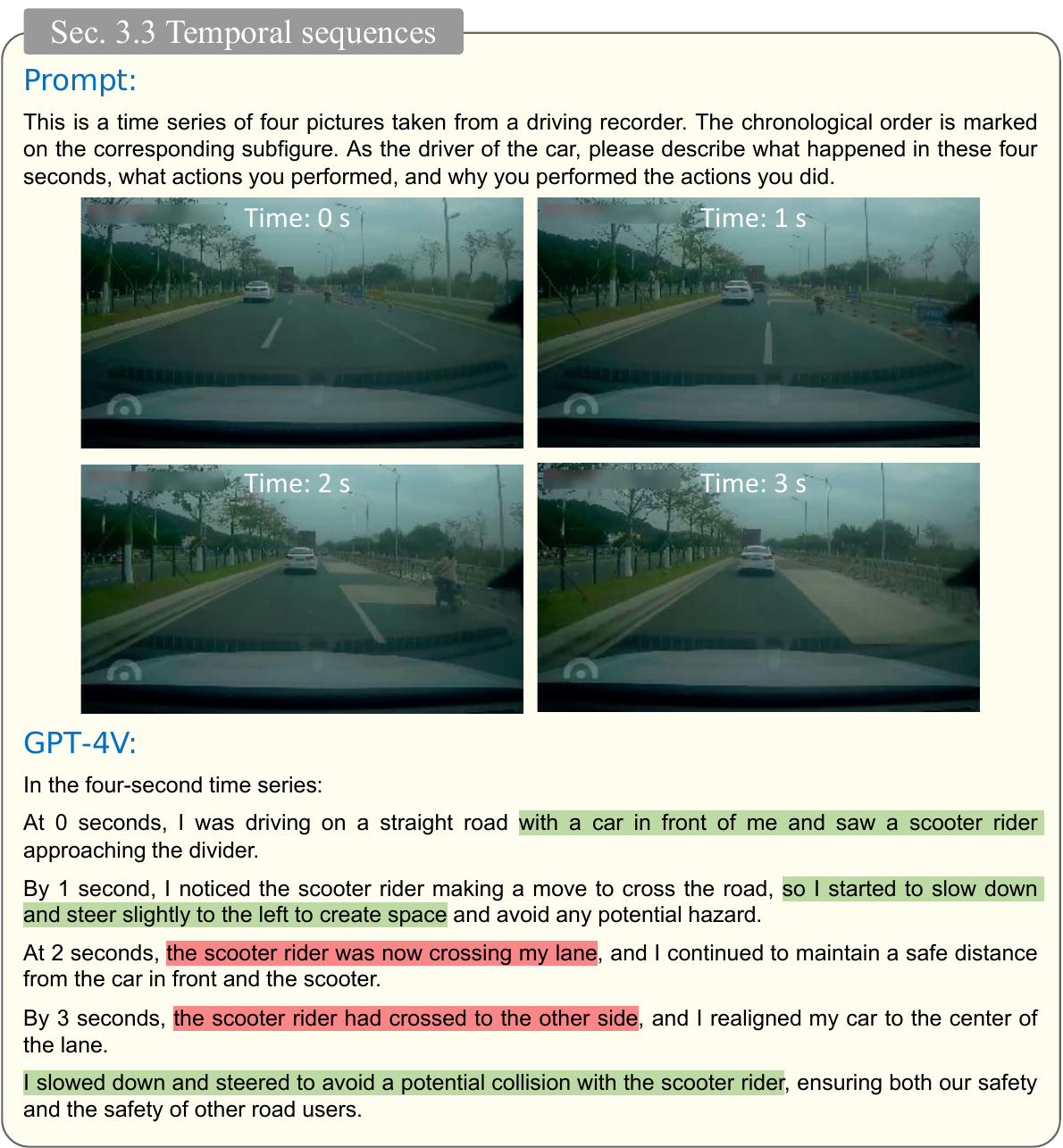}
    \caption[Section~\ref{sec:temporalSequences}: Temporal sequences, Part III]{Illustration of \modelname's ability to understand temporal sequential images.  \colorbox[RGB]{189,218,165}{Green} highlights the right answer in understanding, \colorbox[RGB]{250,156,154}{Red} highlights the wrong answer in understanding. Check Section~\ref{sec:temporalSequences} for detailed discussions.}
    \label{fig: sec3.3 temporal sequences3}
\end{figure*}

However, it's important to note that \modelname is not always able to fully analyze temporal driving scenarios accurately. As illustrated in Figure~\ref{fig: sec3.3 temporal sequences3}, this video segment captures the vehicle making a lane change to overtake the scooter ahead. Regrettably, \modelname incorrectly interprets the scooter's actions as if it were crossing lanes in front of the vehicle and misconstrues the vehicle's behavior as decelerating to yield to the scooter. This underscores the limitations of \modelname in spatial reasoning within temporal video contexts.
Also in Figure~\ref{fig: sec3.3 temporal sequences4}, \modelname once again misidentifies a green light as a red light.

\clearpage
\subsection{Visual-Map Navigation}
\label{sec:visualMapNavigation}

In practical driving scenarios, drivers often utilize auxiliary information from external devices to enhance their decision-making. For instance, a mapping application can provide detailed information about road geometry and route guidance, enabling drivers to make more informed and rational driving decisions. In this section, we equip \modelname with a front-view camera image and corresponding navigation information from the mapping software. This setup allows \modelname to describe the scene and make informed decisions, mirroring the way a human driver would in a similar situation.

\begin{figure*}[htbp]
    \centering
    \includegraphics[width=\textwidth]{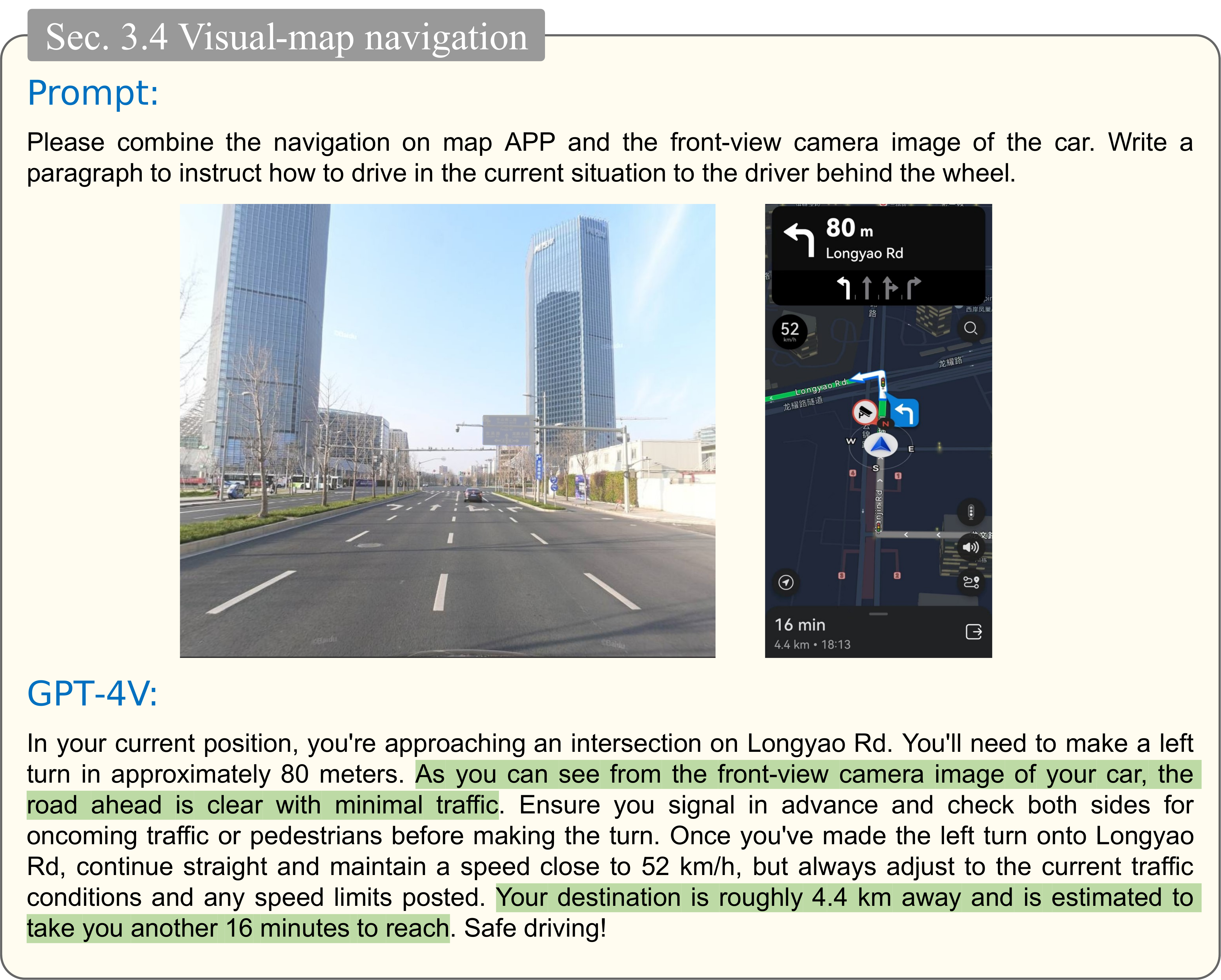}
    \caption[Section~\ref{sec:visualMapNavigation}: Visual-map navigation, Part I]{Illustration of \modelname's ability to get information combines the navigation on map APP and the front-view camera image. \colorbox[RGB]{189,218,165}{Green} highlights the right answer in understanding. Check Section~\ref{sec:visualMapNavigation} for detailed discussions.}
    \label{fig: sec3.4 visual-map navigation1}
\end{figure*}

As depicted in Figure~\ref{fig: sec3.4 visual-map navigation1}, \modelname accurately localizes its position using the front-view camera and map APP information, subsequently executing the correct left-turn maneuver. Utilizing the front view camera, it made reasonable assessments of the road conditions and, in conjunction with the speed information provided by the mapping software, offered appropriate driving suggestions.

\begin{figure*}[htbp]
    \centering
    \includegraphics[width=\textwidth]{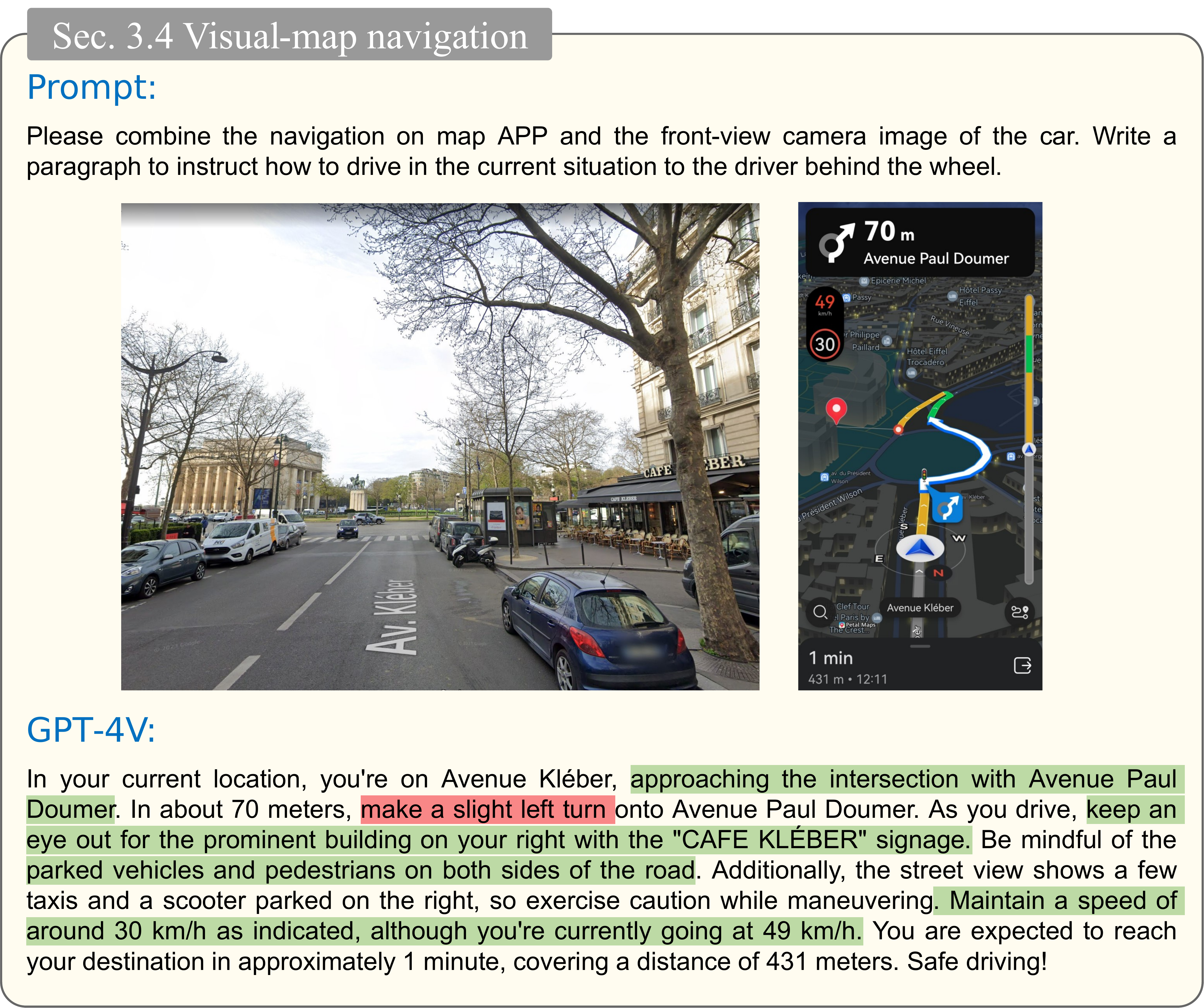}
    \caption[Section~\ref{sec:visualMapNavigation}: Visual-map navigation, Part II]{Illustration of \modelname's ability to get information combines the navigation on map APP and the front-view camera image. \colorbox[RGB]{189,218,165}{Green} highlights the right answer in understanding. Check Section~\ref{sec:visualMapNavigation} for detailed discussions.}
    \label{fig: sec3.4 visual-map navigation2}
\end{figure*}

As illustrated in Figure~\ref{fig: sec3.4 visual-map navigation2}, \modelname is able to accurately localize itself even in a more complex scenario. However, in this particular situation, it incorrectly decided to make a left turn. Despite this, \modelname demonstrated proficiency in recognizing information about parked cars and stores on the roadside based on the front view camera. It also correctly obtained speed and distance information from the mapping software.

%% file: sections/driver-agent.tex
\clearpage
\section{Act as A Driver}
\label{sec:driverAgent}

The ultimate goal of autonomous driving algorithms is to replicate the decision-making abilities of human drivers. Achieving this goal necessitates precise identification, spatial awareness, and an in-depth understanding of spatiotemporal relationships among various traffic elements. In this section, we assess \modelname's full potential in autonomous driving by testing its decision-making prowess across five distinct real-world driving scenarios. These scenarios encompass varying traffic conditions, different times of the day, and multiple driving tasks. During the assessment, ego-vehicle speed and other relevant information are provided, and \modelname is desired to produce the observation and driving actions. Through these carefully designed evaluations, our goal is to push the boundaries of \modelname's capabilities in real-world driving scenarios, shedding light on its potential as a driving force in the future of autonomous transportation.

\subsection{Driving in Parking Lot}
\label{sec:parkinglot}
\begin{figure*}[htbp]
    \centering
    \includegraphics[width=\textwidth]{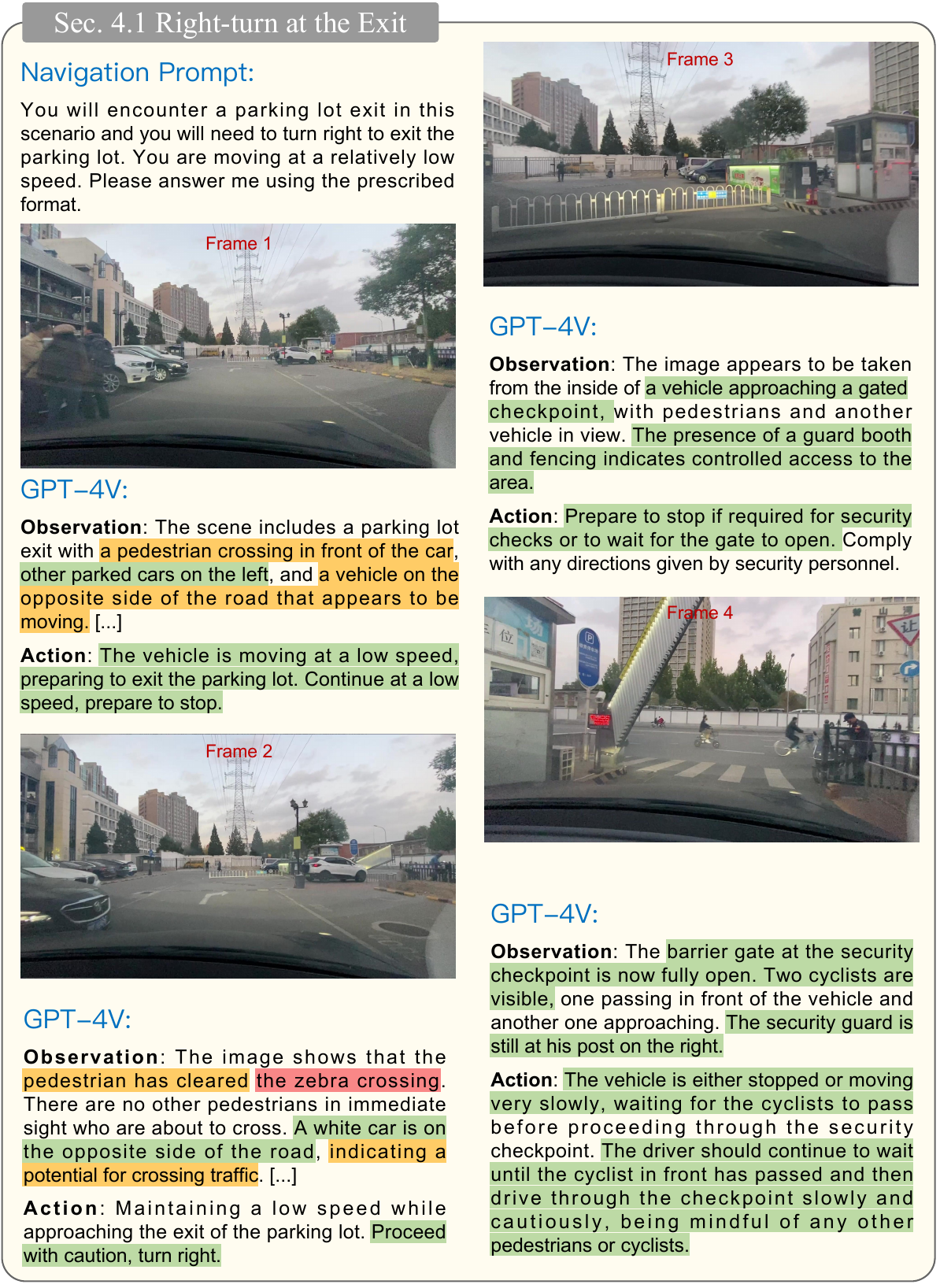}
    \caption[Section~\ref{sec:parkinglot}: Driving in parking lot]{Illustration of \modelname's ability to drive in parking lot. \colorbox[RGB]{189,218,165}{Green} highlights the right answer in understanding. \colorbox[RGB]{250,156,154}{Red} highlights the wrong answer in understanding. \colorbox[RGB]{255, 204, 102}{Yellow} highlights the incompetence in performing the task. Check Section~\ref{sec:parkinglot} for detailed discussions.}
    \label{fig: sec4.1 parkinglot}
\end{figure*}

In this section, we test the driving decision-making ability of \modelname in an enclosed area. The selected scenario is turning right to exit a parking lot, which requires passing through a security check. As shown in Figure~\ref{fig: sec4.1 parkinglot}, in the first frame, \modelname accurately identifies key elements affecting driving, such as pedestrians and vehicle lights. However, \modelname has ambiguity regarding the status of pedestrians and distant vehicles. As a result, it provides conservative driving decisions by maintaining low speed and being prepared to stop. In the second frame, \modelname detects that pedestrians have already left but mistakenly mentions the information of zebra crossings. It still follows a cautious right-turn driving strategy. In the third frame, \modelname accurately recognizes elements such as gated checkpoints, guard booths, and fencing, inferring that the vehicle is approaching the exit and preparing to stop for a security check. In the fourth frame, \modelname correctly identifies that the security checkpoint is now fully open so we can safely exit the parking lot. Additionally, \modelname also locates pedestrians near the exit area and advises waiting for them to pass safely before slowly proceeding out.

From this example,  \modelname can accurately identify key elements within enclosed areas (such as parking lots), including gated checkpoints, guard booths, and fencing. Moreover, \modelname understands driving procedures for leaving parking lots, which requires waiting for security checks and paying attention to pedestrians and vehicles. However, some misjudgments may still occur, such as mentioning zebra crossings erroneously.

\subsection{Turning at Traffic Intersection}
\label{sec:turning_intersection}
\begin{figure*}[htbp]
    \centering
    \includegraphics[width=\textwidth]{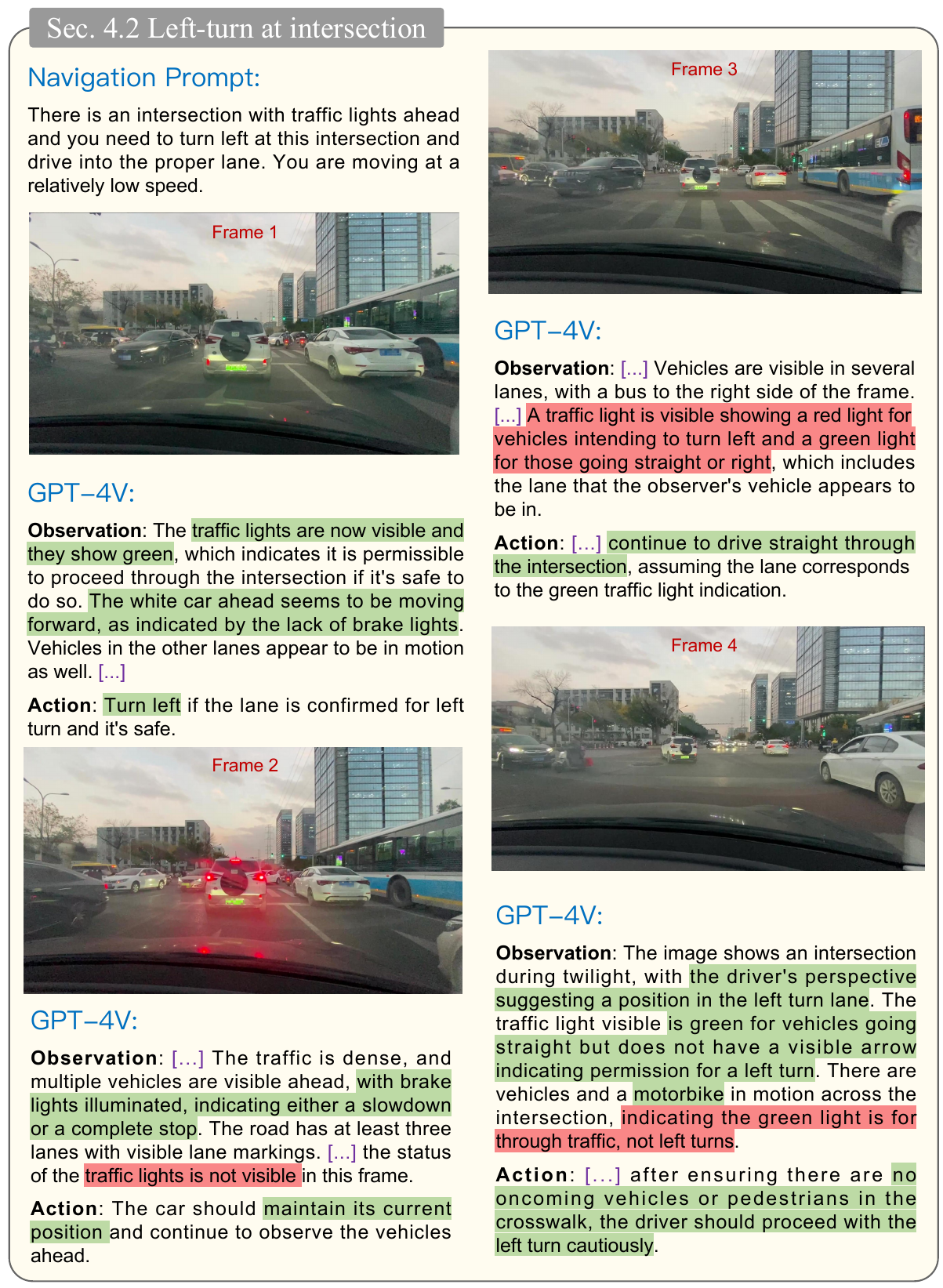}
    \caption[Section~\ref{sec:turning_intersection}: Turning at traffic intersection]{Illustration of \modelname's ability to turn at traffic intersection. \colorbox[RGB]{189,218,165}{Green} highlights the right answer in understanding. \colorbox[RGB]{250,156,154}{Red} highlights the wrong answer in understanding. Check Section~\ref{sec:turning_intersection} for detailed discussions.}
    \label{fig: sec4.2 turning_intersection}
\end{figure*}

In this section, we assess \modelname the turning ability at traffic intersections. As depicted in Figure~\ref{fig: sec4.2 turning_intersection}, the selected scenario is a crossroad with heavy traffic. In the first frame, \modelname observes that the traffic light is green and infers the driving action as continuing to turn left. In the second frame, due to the distance and limited perception fields, \modelname regards that the traffic light is invisible, but it observes that front vehicles were braking based on their taillights. Therefore, its driving strategy was to maintain the current position. In the third frame, \modelname mistakes the status of the traffic light, and deems that turning is not allowed. In the fourth frame, \modelname still mistakes the traffic light status. The final decision is to make a cautious left turn while ensuring safety by avoiding collisions with other vehicles and pedestrians.

This example shows that when making turns at intersections, \modelname pays attention to various information such as traffic lights, and taillights of other vehicles. However, \modelname's ability to identify states of small objects at long distances (such as distant traffic lights) is poor which may affect its behavioral decisions at intersections.

\subsection{Turning at Highway Ramp}
\label{sec:turning_ramp}

\begin{figure*}[htbp]
    \centering
    \includegraphics[width=\textwidth]{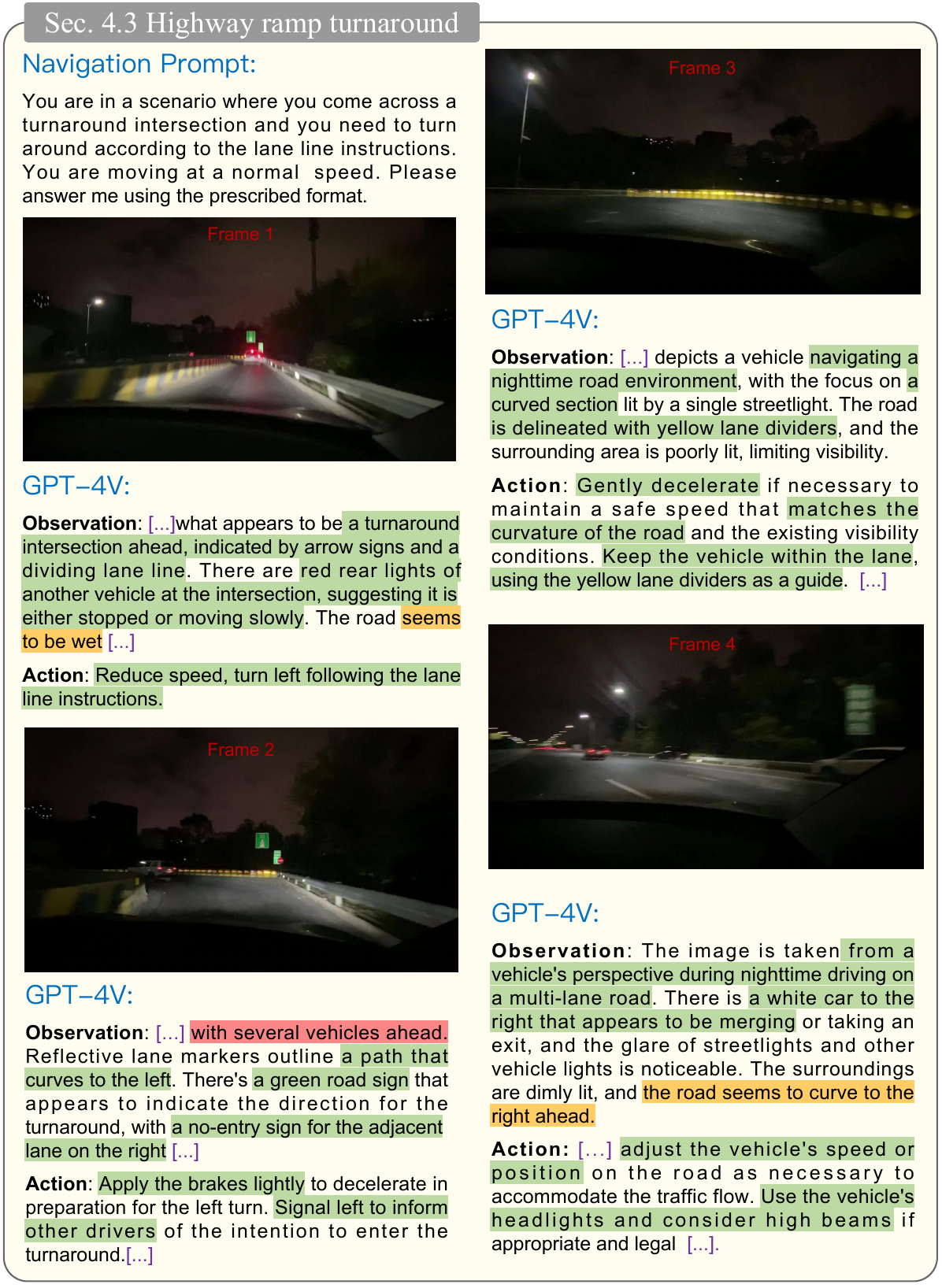}
    \caption[Section~\ref{sec:turning_ramp}: Turning at highway ramp]{Illustration of \modelname's ability to turn at highway ramp. \colorbox[RGB]{189,218,165}{Green} highlights the right answer in understanding. \colorbox[RGB]{250,156,154}{Red} highlights the wrong answer in understanding. \colorbox[RGB]{255, 204, 102}{Yellow} highlights the incompetence in performing the task. Check Section~\ref{sec:turning_ramp} for detailed discussions.}
    \label{fig: sec4.3 turning_ramp}
\end{figure*}

In this section, we test \modelname the capability to drive in highway areas. As illustrated in Figure~\ref{fig: sec4.3 turning_ramp}, we select a challenging scenario where the vehicle needs to perform a highway ramp turnaround at night. In the first frame, \modelname accurately identifies arrow signs and a dividing lane line, and infers from the red taillights of the preceding vehicle that it is slowing down. Therefore, the ego-vehicle should decelerate and follow the lane line. In the second frame, although \modelname mistakes the number of preceding vehicles, it precisely located the lane line and road sign, indicating a left turn ahead. As a result, \modelname suggests applying the brakes lightly and signaling left to inform other drivers. In the third frame, due to limited visibility at night, \modelname only locates yellow lane dividers. Thus, it advises slow driving within the lane lines using these dividers as references. In the fourth frame, \modelname accurately determines that the ego-vehicle has entered the main highway road and observed potential merging vehicles on its right side. Consequently, it decides to adjust speed for highway driving while occasionally activating high beams within legal limits to expand the nighttime visibility range.

From this example, we can see that when driving in highway areas, \modelname follows road signs and assists in decision-making based on the status of surrounding vehicles. However, it has limitations in object recognition and positioning during nighttime.

\subsection{Road Merging}
\label{sec:roadmerging}
\begin{figure*}[htbp]
    \centering
    \includegraphics[width=\textwidth]{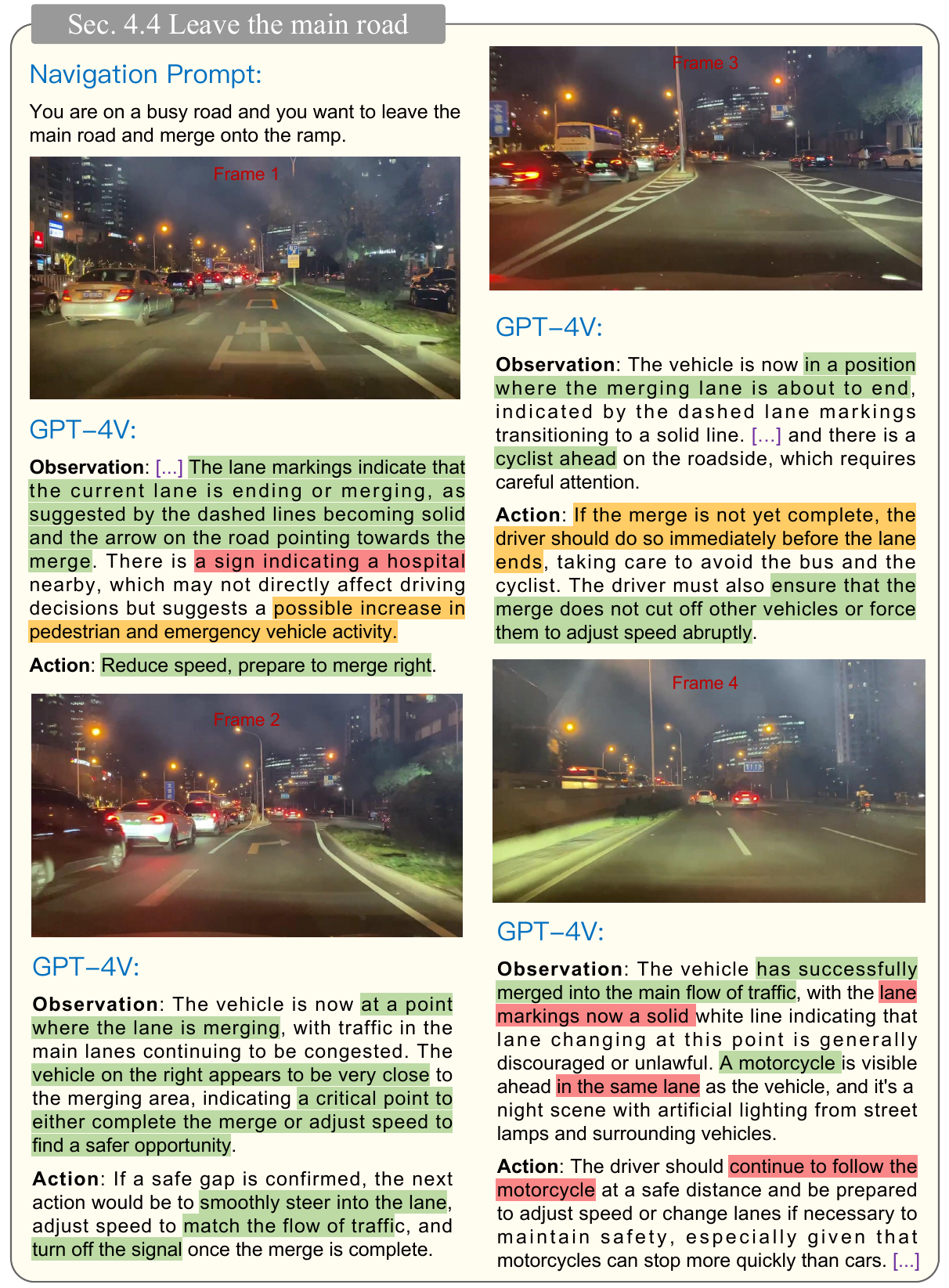}
    \caption[Section~\ref{sec:roadmerging}: Road merging]{Illustration of \modelname's ability to merge onto other road. \colorbox[RGB]{189,218,165}{Green} highlights the right answer in understanding. \colorbox[RGB]{250,156,154}{Red} highlights the wrong answer in understanding. \colorbox[RGB]{255, 204, 102}{Yellow} highlights the incompetence in performing the task. Check Section~\ref{sec:roadmerging} for detailed discussions.}
    \label{fig: sec4.4 roadmerging}
\end{figure*}

In this section, we evaluate the lane merging capability of \modelname. As shown in Figure~\ref{fig: sec4.4 roadmerging}, the selected scenario is exiting the main road at night and merging onto a ramp. In the first frame, \modelname accurately identifies the lane markings and determines that the current lane is ending or merging. Therefore, it decides to decelerate and prepare to merge into the right-turn lane. During this process, it mistakenly recognizes a nearby hospital sign and cautiously considers paying attention to pedestrians and emergency vehicles in the vicinity.
In the second frame, \modelname correctly identifies the merging point and advises smoothly steering into the lane. In the third frame, based on changes in lanes, \modelname predicts that merging is about to end while reminding us to be cautious of other vehicles cutting in.
In the fourth frame, \modelname determines that it has successfully merged onto the road. However, it incorrectly detects a solid white line, and mistakenly believes that a motorcycle is on the same lane. The final decision given was to pay attention to motorcycles on the main road and adjust speed or change lanes if necessary.

From this example, it is observed that \modelname can assess current merge progress by observing changes in lanes and providing reasonable driving suggestions. However, there is still an increased probability of misjudging road signs and lanes during nighttime. Overall, \modelname tends to adopt a conservative approach when it comes to lane merging.

\subsection{U-Turning at Traffic Intersection}
\label{sec:u_turn}

\begin{figure*}[htbp]
    \centering
    \includegraphics[width=\textwidth]{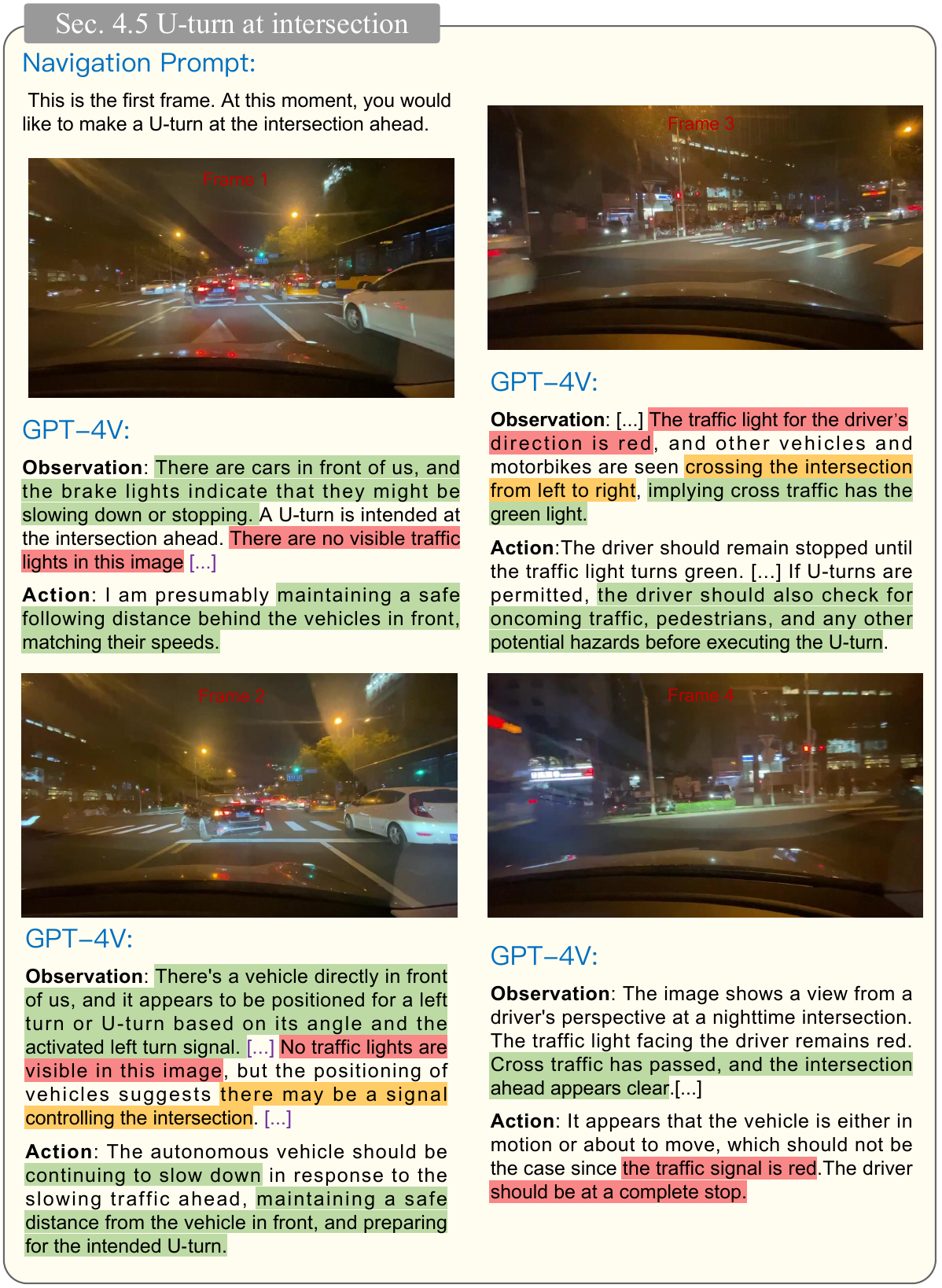}
    \caption[Section~\ref{sec:u_turn}: U-turning at traffic intersection]{Illustration of \modelname's ability to u-turn at traffic intersection. \colorbox[RGB]{189,218,165}{Green} highlights the right answer in understanding. \colorbox[RGB]{250,156,154}{Red} highlights the wrong answer in understanding. \colorbox[RGB]{255, 204, 102}{Yellow} highlights the incompetence in performing the task. Check Section~\ref{sec:u_turn} for detailed discussions.}
    \label{fig: sec4.5 u_turn}
\end{figure*}

In this section, we test the U-turn capability of GP-4V. As depicted in Figure~\ref{fig: sec4.5 u_turn}, we select a scenario where the U-turn is performed at an intersection with heavy nighttime traffic. In the first frame, \modelname accurately identifies other vehicles ahead and reminds ego car to maintain distance, but it omits the distant traffic light. In the second frame, \modelname still fails to locate the traffic light but infers from surrounding vehicle behavior that there might be a signal controlling the intersection. It suggests slowing down to prepare for entering the U-turn. In the third frame, \modelname disregards temporal and spatial context, and mistakes traffic lights on a side road as control signals for its current lane. Therefore, it decides to remain stopped. In the fourth frame, \modelname still mistakes the traffic lights, resulting in maintaining a stationary position as its strategy. 

From this example, we can see that when encountering significant changes in road structure like U-turns, \modelname tends to overlook temporal and spatial contextual relationships. Nevertheless, the overall driving strategies provided are quite conservative.

Through the aforementioned five tests, it is observed that \modelname has initially acquired decision-making abilities similar to human drivers. It can combine the states of various traffic elements (\eg, pedestrians, vehicles, traffic lights, road signs, lanes) to provide the final driving strategy. Besides, \modelname can make reasonable decisions in diverse driving scenarios such as parking lots, intersections, highways, and ramps. Overall, \modelname demonstrates strong adherence to rules and safety awareness with relatively conservative driving strategies. However, there are still limitations in its driving performance. For instance, it struggles to determine the status of distant objects (vehicles, traffic lights), and its perception range is restricted during nighttime. These limitations affect further driving decisions. Additionally, \modelname's ability for spatiotemporal context inference remains limited (\eg, \modelname gets confused in a U-turn scenario with multiple traffic lights).

%% file: sections/limitations-conclusions.tex
\clearpage
\section{Conclusions}

\subsection{Capabilities of \modelname in Autonomous Driving}

In this paper, we have conducted a comprehensive and multi-faceted evaluation of the \modelname in various autonomous driving scenarios. The results indicate that \modelname exhibits capabilities that have the potential to surpass those of existing autonomous driving systems in aspects such as scenario understanding, intention recognition, and driving decision-making. 

In corner cases, \modelname leverages its advanced understanding capabilities to handle out-of-distribution scenarios and can accurately assess the intentions of surrounding traffic participants.
\modelname utilizes multi-view images and temporal photos to achieve a complete perception of the environment, accurately identifying dynamic interactions between traffic participants. Moreover, it can infer the underlying motives behind these behaviors. 
As highlighted in Section \ref{sec:driverAgent}, we also witnessed the performance of \modelname in making continuous decisions on open roads. 
It can even interpret the user interface of navigation apps in a human-like manner, assisting and guiding drivers in their decision-making processes.

Overall, the performance of \modelname demonstrates the significant potential of Vision-Language Models (VLMs) to tackle complex challenges in the field of autonomous driving.

\subsection{Limitations of \modelname in Autonomous Driving}

However, during our testing, we also found that \modelname performs poorly on the following tasks:

\textbf{Distinguishing left from right:}
As depicted in Figure \ref{fig: sec3.2 multi-view images2}, there were instances where the model struggled with recognizing directions, which is a critical aspect of autonomous navigation. Similar issues are also observed in Figures \ref{fig:sec2.2 front view 2} and \ref{fig: sec3.3 temporal sequences3}. These figures highlight the model’s occasional confusion when interpreting complex junctions or making lane-changing decisions.

\textbf{Traffic light recognition:} Issues are observed in Figures \ref{fig:sec2.2 simulated view}, \ref{fig: sec3.1 corner cases3}, \ref{fig: sec3.3 temporal sequences4}, \ref{fig: sec4.2 turning_intersection} and \ref{fig: sec4.5 u_turn}. We suspect this problem is due to the extensive semantic information contained within the full image, leading to a loss in the embedding information of traffic lights. When the region of the traffic lights in the image is cropped and inputted separately, the model is capable of successful recognition shown in Figure \ref{fig: sec2.1 traffic light understanding 2}.

\textbf{Vision Grounding tasks:} As shown in Figure \ref{fig:sec2.2 front view 1}, \modelname finds it difficult to specify pixel-level coordinates or bounding boxes, managing only to indicate approximate areas within the image.

\textbf{Spatial Reasoning:} 
Accurate spatial reasoning is paramount for the safe operation of autonomous vehicles. Whether it is the stitching of multiview images as illustrated in Figure \ref{fig: sec3.2 multi-view images3} or the estimation of the relative positional relationship between a scooter and the self-driving car as shown in Figure \ref{fig: sec3.3 temporal sequences3}, \modelname struggles with making precise judgments. This may stem from the inherent complexity in understanding and interpreting three-dimensional space based on two-dimensional image inputs.

Additionally, issues were found with the model's interpretation of non-English traffic signs, which poses a challenge in regions where multiple languages are used on signage. The accuracy of counting traffic participants was also found to be less reliable in congested environments where overlapping objects can occur.


In conclusion, the above limitations indicate that even the most advanced Vision-Language Models (VLMs) currently exhibit deficiencies in basic directional recognition and traffic light identification, as well as a lack of 3D spatial reasoning capabilities. 
Furthermore, VLMs struggle to accurately localize key entities in various scenarios, suggesting that they are not yet suitable replacements for the perception methods used in existing autonomous driving pipelines.
However,  it is noteworthy that VLMs demonstrate a deep understanding of traffic common sense and strong generalization capabilities in out-of-distribution cases. 
Looking ahead, a key area of development will be to integrate the innate common sense knowledge of VLMs with conventional autonomous driving perception techniques. In addition, ensuring the safety and reliability of VLM outputs remains an essential and ongoing challenge.

%% file: main.bbl
\begin{thebibliography}{10}

\bibitem{TSDD}
Chinese traffic sign database.
\newblock \url{http://www.nlpr.ia.ac.cn/pal/trafficdata/detection.html}.

\bibitem{BaoMM2020}
Wentao Bao, Qi~Yu, and Yu~Kong.
\newblock Uncertainty-based traffic accident anticipation with spatio-temporal relational learning.
\newblock In {\em ACM Multimedia Conference}, May 2020.

\bibitem{caesar2020nuscenes}
Holger Caesar, Varun Bankiti, Alex~H Lang, Sourabh Vora, Venice~Erin Liong, Qiang Xu, Anush Krishnan, Yu~Pan, Giancarlo Baldan, and Oscar Beijbom.
\newblock nuscenes: A multimodal dataset for autonomous driving.
\newblock In {\em Proceedings of the IEEE/CVF conference on computer vision and pattern recognition}, pages 11621--11631, 2020.

\bibitem{che2019d}
Zhengping Che, Guangyu Li, Tracy Li, Bo~Jiang, Xuefeng Shi, Xinsheng Zhang, Ying Lu, Guobin Wu, Yan Liu, and Jieping Ye.
\newblock D$^2$-city: A large-scale dashcam video dataset of diverse traffic scenarios.
\newblock 2019.

\bibitem{chen2023driving}
Long Chen, Oleg Sinavski, Jan H{\"u}nermann, Alice Karnsund, Andrew~James Willmott, Danny Birch, Daniel Maund, and Jamie Shotton.
\newblock Driving with llms: Fusing object-level vector modality for explainable autonomous driving.
\newblock {\em arXiv preprint arXiv:2310.01957}, 2023.

\bibitem{dosovitskiy2017carla}
Alexey Dosovitskiy, German Ros, Felipe Codevilla, Antonio Lopez, and Vladlen Koltun.
\newblock Carla: An open urban driving simulator.
\newblock In {\em Conference on robot learning}, pages 1--16. PMLR, 2017.

\bibitem{du2022glm}
Zhengxiao Du, Yujie Qian, Xiao Liu, Ming Ding, Jiezhong Qiu, Zhilin Yang, and Jie Tang.
\newblock Glm: General language model pretraining with autoregressive blank infilling.
\newblock In {\em Proceedings of the 60th Annual Meeting of the Association for Computational Linguistics (Volume 1: Long Papers)}, pages 320--335, 2022.

\bibitem{fu2023drive}
Daocheng Fu, Xin Li, Licheng Wen, Min Dou, Pinlong Cai, Botian Shi, and Yu~Qiao.
\newblock Drive like a human: Rethinking autonomous driving with large language models.
\newblock {\em arXiv preprint arXiv:2307.07162}, 2023.

\bibitem{kim2018textual}
Jinkyu Kim, Anna Rohrbach, Trevor Darrell, John Canny, and Zeynep Akata.
\newblock Textual explanations for self-driving vehicles.
\newblock {\em Proceedings of the European Conference on Computer Vision (ECCV)}, 2018.

\bibitem{li2022coda}
Kaican Li, Kai Chen, Haoyu Wang, Lanqing Hong, Chaoqiang Ye, Jianhua Han, Yukuai Chen, Wei Zhang, Chunjing Xu, Dit-Yan Yeung, et~al.
\newblock Coda: A real-world road corner case dataset for object detection in autonomous driving.
\newblock In {\em European Conference on Computer Vision}, pages 406--423. Springer, 2022.

\bibitem{mao2023gpt}
Jiageng Mao, Yuxi Qian, Hang Zhao, and Yue Wang.
\newblock Gpt-driver: Learning to drive with gpt.
\newblock {\em arXiv preprint arXiv:2310.01415}, 2023.

\bibitem{gpt3.5}
OpenAI.
\newblock \url{https://chat.openai.com}, 2023.

\bibitem{gpt4v_3}
OpenAI.
\newblock Chatgpt can now see, hear, and speak.
\newblock \url{https://openai.com/blog/chatgpt-can-now-see-hear-and-speak}, 2023.

\bibitem{gpt4}
OpenAI.
\newblock Gpt-4 technical report, 2023.

\bibitem{gpt4v}
OpenAI.
\newblock Gpt-4v(ision) system card.
\newblock 2023.

\bibitem{gpt4v_2}
OpenAI.
\newblock Gpt-4v(ision) technical work and authors.
\newblock \url{https://openai.com/contributions/gpt-4v}, 2023.

\bibitem{sun2020scalability}
Pei Sun, Henrik Kretzschmar, Xerxes Dotiwalla, Aurelien Chouard, Vijaysai Patnaik, Paul Tsui, James Guo, Yin Zhou, Yuning Chai, Benjamin Caine, et~al.
\newblock Scalability in perception for autonomous driving: Waymo open dataset.
\newblock In {\em Proceedings of the IEEE/CVF conference on computer vision and pattern recognition}, pages 2446--2454, 2020.

\bibitem{touvron2023llama}
Hugo Touvron, Thibaut Lavril, Gautier Izacard, Xavier Martinet, Marie-Anne Lachaux, Timoth{\'e}e Lacroix, Baptiste Rozi{\`e}re, Naman Goyal, Eric Hambro, Faisal Azhar, et~al.
\newblock Llama: Open and efficient foundation language models.
\newblock {\em arXiv preprint arXiv:2302.13971}, 2023.

\bibitem{touvron2023llama2}
Hugo Touvron, Louis Martin, Kevin Stone, Peter Albert, Amjad Almahairi, Yasmine Babaei, Nikolay Bashlykov, Soumya Batra, Prajjwal Bhargava, Shruti Bhosale, et~al.
\newblock Llama 2: Open foundation and fine-tuned chat models.
\newblock {\em arXiv preprint arXiv:2307.09288}, 2023.

\bibitem{wen2023dilu}
Licheng Wen, Daocheng Fu, Xin Li, Xinyu Cai, Tao Ma, Pinlong Cai, Min Dou, Botian Shi, Liang He, and Yu~Qiao.
\newblock Dilu: A knowledge-driven approach to autonomous driving with large language models.
\newblock {\em arXiv preprint arXiv:2309.16292}, 2023.

\bibitem{wu2023add}
Zizhang Wu, Xinyuan Chen, Hongyang Wei, Fan Song, and Tianhao Xu.
\newblock Add: An automatic desensitization fisheye dataset for autonomous driving.
\newblock {\em Engineering Applications of Artificial Intelligence}, 126:106766, 2023.

\bibitem{yang2023dawn}
Zhengyuan Yang, Linjie Li, Kevin Lin, Jianfeng Wang, Chung-Ching Lin, Zicheng Liu, and Lijuan Wang.
\newblock The dawn of lmms: Preliminary explorations with gpt-4v (ision).
\newblock {\em arXiv preprint arXiv:2309.17421}, 9, 2023.

\bibitem{yu2022dair}
Haibao Yu, Yizhen Luo, Mao Shu, Yiyi Huo, Zebang Yang, Yifeng Shi, Zhenglong Guo, Hanyu Li, Xing Hu, Jirui Yuan, et~al.
\newblock Dair-v2x: A large-scale dataset for vehicle-infrastructure cooperative 3d object detection.
\newblock In {\em Proceedings of the IEEE/CVF Conference on Computer Vision and Pattern Recognition}, pages 21361--21370, 2022.

\bibitem{zeng2022glm}
Aohan Zeng, Xiao Liu, Zhengxiao Du, Zihan Wang, Hanyu Lai, Ming Ding, Zhuoyi Yang, Yifan Xu, Wendi Zheng, Xiao Xia, et~al.
\newblock Glm-130b: An open bilingual pre-trained model.
\newblock {\em arXiv preprint arXiv:2210.02414}, 2022.

\bibitem{zheng2022citysim}
Ou~Zheng, Mohamed Abdel-Aty, Lishengsa Yue, Amr Abdelraouf, Zijin Wang, and Nada Mahmoud.
\newblock Citysim: A drone-based vehicle trajectory dataset for safety oriented research and digital twins.
\newblock {\em arXiv preprint arXiv:2208.11036}, 2022.

\end{thebibliography}
